\UseRawInputEncoding
\documentclass[lettersize,journal]{IEEEtran}
\usepackage{lineno}
\usepackage{epsfig}
\usepackage{chngpage}
\usepackage{float}
\usepackage{amsmath}
\usepackage{graphicx}
\usepackage{array}
\usepackage{diagbox}
\usepackage{epstopdf}
\usepackage{algorithm}
\usepackage{algorithmicx}
\usepackage{algpseudocode}
\usepackage{threeparttable}
\usepackage{subfigure}
\usepackage{multirow}
\usepackage{longtable}
\usepackage{bm}
\usepackage[table,xcdraw]{xcolor}
\usepackage{url}
\usepackage{booktabs}
\usepackage{multirow}
\usepackage[colorlinks,linkcolor=blue]{hyperref}
\usepackage{algorithm}
\usepackage{algorithmicx}
\usepackage{algpseudocode}
\usepackage{amsmath}
\usepackage{amssymb}%空集符号
\newtheorem{myDef}{Definition}

\begin{document}

\title{Deep Graph Learning for Anomalous Citation Detection}

        % <-this % stops a space
\author{Jiaying Liu, Feng Xia,~\IEEEmembership{Senior Member,~IEEE}, Xu Feng, Jing Ren, and Huan Liu,~\IEEEmembership{Fellow,~IEEE}
% <-this % stops a space
\thanks{J. Liu is with School of Economics and Management, Dalian University of Technology, Dalian 116024, China.}% <-this % stops a space
\thanks{F. Xia and J. Ren are with School of Engineering, IT and Physical Sciences, Federation University Australia, Ballarat, VIC 3353, Australia.}
\thanks{X. Feng is with School of Software, Dalian University of Technology, Dalian 116620, China.}
\thanks{H. Liu is with School of Computing, Informatics, and Decision Systems Engineering, Arizona State University, Tempe, AZ 85281, USA.}
\thanks{Corresponding author: Feng Xia; email: f.xia@ieee.org.}}

\markboth{IEEE Transactions on IEEE Transactions on Neural Networks and Learning Systems, ~Vol.~0, No.~0, ~2022}%
{J. Liu  \MakeLowercase{\textit{et al.}}: Deep Graph Learning for Anomalous Citation Detection}

\IEEEpubid{0000--0000/00\$00.00~\copyright~2021 IEEE}

\maketitle

\begin{abstract}
Anomaly detection is one of the most active research areas in various critical domains, such as healthcare, fintech, and public security. However, little attention has been paid to scholarly data, i.e., anomaly detection in a citation network. Citation is considered as one of the most crucial metrics to evaluate the impact of scientific research, which may be gamed in multiple ways. Therefore, anomaly detection in citation networks is of significant importance to identify manipulation and inflation of citations. To address this open issue, we propose a novel deep graph learning model, namely GLAD (Graph Learning for Anomaly Detection), to identify anomalies in citation networks. GLAD incorporates text semantic mining to network representation learning by adding both node attributes and link attributes via graph neural networks. It exploits not only the relevance of citation contents but also hidden relationships between papers. Within the GLAD framework, we propose an algorithm called CPU (Citation PUrpose) to discover the purpose of citation based on citation texts. The performance of GLAD is validated through a simulated anomalous citation dataset. Experimental results demonstrate the effectiveness of GLAD on the anomalous citation detection task.
\end{abstract}


\section{Introduction}
\IEEEPARstart{A}{n} objective and fair evaluation of the impact of published research is essential for science itself. Since citations are important for assessing the scholarly impact of academic entities (e.g., publications, scholars, and institutions), it becomes essential to avoid making citation-based bibliometrics strongly flawed and defective. A branch of Science of Science (SciSci)~\cite{fortunato2018science} entitled Scholarly Network Analysis (SNA)~\cite{kong2019academic} has emerged intending to discover meaningful insights to implement data-driven research decisions. Citation network analysis is one of the important points of interest in SNA, aiming to find relationships between cited papers and citing papers. Expansion of large open databases such as Web of Science (WOS), Scopus, and Microsoft Academic Graph (MAG) has opened up opportunities for multi-perspective and systematic analysis of citation networks~\cite{Wang2020TWEB}.

%The fast-growing scholarly big data promotes the development of data mining techniques, leading to making relevant analysis consistently in-depth, which not only helps to reveal the law of science itself but also creates immeasurable value.
%%现有工作忽略了异常的引用，讲什么是异常引用，它有多重要，在很多方面有体现 举例子

Traditionally, the quantity of citations is one measure of the quality of a paper, or the impact of academic entities~\cite{Bai2020IACCESS}. For example, existing bibliometric indicators such as h-index, g-index, and Journal Impact Factor (JIF) all focus on quantitative evaluation aspect while ignoring qualitative aspects. However, with the explosive growth of publications, the number of citations increases fast. As citation-based metrics are transformed into ``academic currency", the credit of citations also implies economical gains. Suspicious citations are one of many other issues in the context of broader unethical practices that affect science academic integrity and fairness. Specifically, some citations are solely intended to boost the unwarranted impact of the publication, rather than disseminate scientific advancements. Although researchers have realized that not all citations are equal and tried to assign different weights to citations with different types, most studies have focused on distinguishing and weighting citations of a single type. In fact, as Prabha~\cite{prabha1983some} illustrated, more than 2/3 references in a paper are unnecessary, which also confirms the existence of dubious citations.

%%异常的引用的危害
These misused citations can cause a series of side effects. On the one hand, they affect citation-based measurement indices related to scholar promotion and reward in bibliometrics. On the other hand, they may mislead readers about the paper correct value. Worst of all, a large number of meaningless citations can negatively impact academic integrity. Therefore, anomaly detection in the citation network is critical to identify manipulation and inflation of citations. To the best of our knowledge, there exists no related method in the relevant literature that is directly applicable to automatic detection of anomalous citations.

Conventional anomaly detection techniques cannot tackle this problem well because of the complexity of graph data (e.g., irregular structures and relational dependencies). Compared with the traditional conventional detection methods, anomaly detection methods based on graph learning can preserve the node attributes and network structures at the same time during the learning process. However, most of the anomaly methods based on graph learning focus on detecting anomalous nodes in the network, such as GCNSI~\cite{dong2019multiple} and GCAN~\cite{lu2020gcan}. To date, only a few works have put their efforts into utilizing graph learning techniques for anomalous edge detection. At the same time, most of the anomalous edge detection methods are task-dependent and are designed for specific tasks such as spam review detection and rumor detection~\cite{bian2020rumor,li2019spam,Yu2020TCSSdetecting}. These methods cannot be directly applied to anomalous citation detection. The major issue of direct usage of graph learning in anomalous citation detection lies in the fact that they can't integrate edge feature learning process and semantic analysis such as citation purpose, which cannot be ignored in the learning process. It drives us to propose a new method that considers both anomalous citation features and the structure of the citation network to identify anomalous citations.

In this work, as a first step, we define a novel class of citations, namely anomalous citations, which contains two aspects: semantic information and relational information. We say that a citation is anomalous if i) the cited paper and the citing paper are disparate from the perspective of contents, and there is no clear purpose in the scientific citation text, and ii) the citing paper contains excessive relational citations. We then propose a framework based on deep graph learning, namely GLAD (Graph Learning for Anomaly Detection), to detect the anomalous paper pairs. GLAD incorporates text semantic mining to network representation learning by adding both node attributes and link attributes via graph neural networks. In addition, we propose an algorithm called CPU (Citation PUrpose) to identify the purpose of citation based on citation context. To validate the performance of GLAD, we generate a simulated dataset for anomalous citations. We compare GLAD's performance with that of state-of-the-art approaches for network embedding. GLAD outperforms baseline approaches by up to 37\% in the anomalous citation detection task. Our major contributions can be summarized as follows:

\begin{itemize}
  \item We formally define a novel class of relationships introducing both semantic information and relational information, which covers universal real-world issues.
  \item We provide an effective algorithm named CPU based on transfer learning to solve the problem of unmarked citation purposes. It can automatically judge the relationship between the cited paper and the citing paper based on the citation context.
  \item We propose a deep graph learning framework, namely GLAD, for anomalous citation detection. GLAD can model rich information with integrating attributes such as text content and author relationship.
  \item Extensive experiments have been conducted. The results verify the effectiveness and superiority of GLAD as compared against state-of-the-art baselines.
\end{itemize}

The rest of this paper proceeds as follows. Section~\ref{sec:Related work} briefly reviews the related work the related work of SNA and GNNs. In Section~\ref{sec:Problem Definition}, we introduce mathematical preliminaries and formally formulate the problem of anomalous citation detection. Section~\ref{sec:Framework} illustrates our optimal solution, GLAD and describes its architecture in detail. Experimental settings and results are described in Section~\ref{sec:Experiments}. Finally in Section~\ref{sec:Conclusion and Future Work}, we conclude the paper.

\section{Related Work}
\label{sec:Related work}
Although numerous techniques have been developed for anomalies detection in different areas such as financial security and image/video surveillance~\cite{fiore2019using}, limited work has paid attention to anomalous citation detection in academia. Hence, in this part, we introduce topics that are most relevant to our research topics, including scholarly network analysis and graph neural networks.
\subsection{Scholarly Network Analysis}
%Academic network analysis is a vital task in the era of big scholarly data~\cite{fortunato2018science,xia2017big,Shifu22019tkde}. It aims to reveal inherent laws and quantitative characteristics within science of science by mining different relationships among academic entities (such as journals, papers, and researchers). Techniques for academic network analysis are covered with multiple fields including bibliometrics, natural language processing, computer science, and complex network analysis. In general, academic network analysis is a mature research topic that can be applied in extensive applications from three levels: actor, relationship, and network~\cite{kong2019academic}. Specific applications consist of expert finding~\cite{cifariello2019wiser}, scientific impact evaluation~\cite{wang2013quantifying,bai2016pncoirank}, paper recommendation~\cite{jeong2020context}, etc. In this paper, we devise a semantic-based citation anomaly detection framework. The following briefly emphasizes some related work of anomalous citation.

It is well known that citation count is an important index to measure academic entities such as papers, journals, and scholars~\cite{xia2017big,Kong2020TKDDgene}. Indeed, references in the paper are often overlooked and there is also randomness to a certain extent in citation behavior. Prabha~\cite{prabha1983some} has pointed out that only less than 1/3 references in each paper are mandatory. Moustafa~\cite{moustafa2016aberration} introduces the idea that there are multiple inherent biases in citation practice, which will make citation-based bibliometric measures strongly flawed and defective. For example, authors can easily inflate their h-indices and distort scientific knowledge towards more conformism by manipulating self-citations~\cite{bartneck2011detecting}. Inappropriate co-author self-citations and collaborative self-citations can also mislead and distort scientific literature, thereby challenging scientific fairness. Reciprocal citations indicate that authors prefer to cite publications of people who cite their own work rather than those who do not~\cite{corbyn2010easy}. Moreover, to inflate the journal impact factor, some editors have encouraged authors to cite the journal's own papers, which will lead to journal citation stacking. Previous studies have explored the relationships between journal impact factors and self-citations~\cite{mimouni2016self}. Another similar kind of citations is called coercive induced self-citation~\cite{ioannidis2015generalized}, which may occur in reviewers, grants, advisors, etc.

Under pressure to ensure that journals/authors maintain high-value bibliometric indicators, i.e., JIF and h-index, some editors have artificially generated citations for their journals/papers by forcing authors to submit papers or by writing reviews. ``Citation cartel" refers to the behavior of journal groups exchanging citations with excessive frequency~\cite{franck1999scientific}. Increasing competition in the academic publishing market has led to such practice becoming more common in recent years. Many instances of cartels have been reported. For instance, Clarivate Analytics, the publisher of the annual Journal Citation Report (JCR), has featured and suspended journals which distort JIF caused by different anomalous citation patterns\footnote{http://help.prod-incites.com/incitesLiveJCR/JCRGroup/titleSuppressions}. The Scholarly Kitchen also reports citation cartel cases\footnote{https://scholarlykitchen.sspnet.org/2012/04/10/emergence-of-a-citation-cartel/}. Recipient journals can enhance their JIF by up to 94\% by receiving citations from donor journals in the JIF years. As such indicators can reflect the quality and prestige of journals, it is essential to detect such behaviors for academic fairness.

Unlike self-citations which are easy to detect, the issue of journal citation cartels has not been widely documented. Few algorithms are designed for detecting citation cartels. Determining the boundaries of ``excessive" has become the biggest challenge. It seems feasible to detect such anomalous pairs based on network community approaches. However, such approaches may lead to false positives because publications in one journal tend to cite publications in other similar journals in the same research field, thus forming a closely-connected community.

Although the aforementioned literature explains the existence of anomalous citations through citation network analysis and publishing ethics elaboration, to the best of our knowledge, automatic anomalies detection methods for citation networks have not been well studied. On the one hand, existing graph anomaly detection algorithms cannot capture abnormal subtle signal strengths to support pairwise anomalies detection in citation networks. On the other hand, the lack of a ground-truth dataset with considerable size challenges the application of supervised machine learning methods in anomalies detection.

\subsection{Graph Neural Networks}
The concept of graph neural network (GNN) was initially outlined by Gori et al.~\cite{gori2005new}, which is a kind of neural network specially designed for processing highly irregular graph data. GNN models can be categorized into Recurrent Graph Neural Networks (RecGNNs), Convolutional Graph Neural Networks (ConvGNNs), Graph Autoencoders (GAEs), and Spatial-Temporal Graph Neural Networks (STGNNs)~\cite{Xia2021TAI}.

RecGNNs rely on recurrent neural architectures and apply the same parameter set recurrently over nodes to learn node representations. They assume a node constantly exchanges information/message with its neighbors in a graph until a stable equilibrium is reached. Early research on RecGNNs includes GNN~\cite{scarselli2008graph}, GraphESN~\cite{gallicchio2010graph}, Gated GNN~\cite{li2015gated}. ConvGNNs are closely related to RecGNNs. Different from RecGNNs, ConvGNNs handle the cyclic mutual dependencies architecturally by adopting a fixed number of layers with different weights. ConvGNNs can be categorized into spectral-based ConvGNNs and spatial-based ConvGNNs. Spectral-based approaches define graph convolution based on graph signal processing and spatial-based ConvGNNs including DCNN~~\cite{atwood2016diffusion}, GraphSAGE~\cite{hamilton2017inductive}, FastGCN~\cite{chen2018fastgcn}, GIN~\cite{xu2019powerful}, and SemanticGCN~\cite{wu2021learning} inherit ideas from RecGNNs by information propagation. Recently, scholars have paid attention to compute mutual information between high dimensional input/output pairs of deep neural networks in diverse domains such as images and speech. Specifically, in deep graph learning, Velickovic et al.~\cite{velickovic2019deep} propose Deep Graph Infomax (DGI) for learning representations for graph-structured inputs by maximizing mutual information between global representations and the local patches in an unsupervised manner. Following on DGI, Ren et al.~\cite{ren2020heterogeneous} propose Heterogeneous Deep Graph Infomax (HDGI) to learn node representations by maximizing local-global mutual information.

GAEs learn network embeddings and graph generative distributions by encoding nodes/graphs into the latent space and reconstructing the graph from encoded information. GAEs are available for network embedding (learn latent representations by graph structural information reconstruction)~\cite{tu2018deep,yu2018learning} and graph generation (generate step by step or output the graph all at once)~\cite{ma2018constrained,bojchevski2018netgan}. STGNNs aim to capture the graph dynamics while assuming the interdependence between connected nodes. The dominant merit of STGNNs is that they can capture spatial and temporal dependencies of a graph simultaneously. STGNNs are divided into two directions, i.e., RNN-based methods~\cite{zhang2018gaan} and CNN-based methods~\cite{guo2019attention}. RNN-based methods filter the input and use state passed to the loop unit by graph convolution to learn the spatial-temporal dependence. CNN-based methods interleave 1-D-CNN layers with graph convolutional
layers in a non-recursive manner. Compared with RNN-based methods, CNN-based approaches can realize parallel computing, stable gradients, and low-memory requirements.

In addition to typical GNNs mentioned above, many variations of GNN have been developed recently, i.e., GRCN~\cite{yu2020graph}, GAUGM~\cite{zhao2021data}, GNN-Guard~\cite{zhang2020gnnguard}, and AMGCN~\cite{wang2020gcn}. It has been proven that GNN brings breakthrough improvements to fulfilling various tasks in different areas such as e-commerce, chemistry, and biomedicine. Specific applications of GNNs include computer vision, natural language processing, recommender systems, and so on~\cite{Shifu22019tkde,Wang2020TCSSmodel}.

Graph anomaly detection with graph learning has received growing attention recently. Existing work can be categorized into anomalous nodes detection methods, anomalous edge detection methods, and anomalous subgraphs detection methods according to the anomalous graph objects that they can detect. The detection of anomalous nodes is the focus of scholars. Scholars focus on using GAT, GCN, and GAE to identify malicious users, fake news, and financial fraud. Only few works have put their efforts into the anomalous edge and subgraph detection based on graph learning. Ouyang et al.~\cite{ouyang2020unified} model the distribution of edges based on a neural network and identify existing edges that are least likely to appear as anomalies. AddGraph~\cite{zheng2019addgraph} is a GCN-based framework for anomalous edge detection in dynamic networks. There is a huge gap between the existing anomalous edge/subgraph/graph detection techniques and the emerging demands for more advanced solutions in various applications.

\section{Preliminaries and Definitions}
\label{sec:Problem Definition}
Before we give details about the proposed framework, we first delve into the concept of anomalous citations and mathematical preliminaries used throughout the paper.

\subsection{Definition of Anomalous Citation}
The aforementioned literature explains the existence of different anomalous citation patterns through citation network analysis and publishing ethics elaboration. Unfortunately, there is no clear definition of anomalous citation, as well as automatic anomalous citation detection methods. This research aims to solve the problem of anomalies detection in the citation network. To this end, as a first step, we refer to the above-mentioned aberration citations to give the definition of anomalous citations.

\begin{myDef}\emph{\textbf{Anomalous Citations}}.

In this paper, the citation relationship between paper pairs that satisfies the conditions described below is regarded as anomalous:
\begin{enumerate}
  \item The cited paper and the citing paper are disparate from the perspective of contents, and there is no clear purpose in the scientific citation context; and
  \item The citing paper contains excessive relational citations.
\end{enumerate}

\end{myDef}

\subsection{Mathematical Preliminaries}
In this part, we introduce the notations we will use in the remainder of the paper. An academic citation network $G$ consists of a set of nodes  $V=\{v_1, v_2,...,v_{|V|}\}$ (represent papers) and a set of edges $E=\{e_1, e_2,...,e_{|E|}\}$, which describe the interaction in terms of citing-cited relationship between nodes. The connection between nodes can be described by the adjacency matrix $\bm{A}$. $\bm{A}_{ij}$ denotes the element in $i$-th row and $j$-th column of $\bm{A}$. If there is a link between node $v_i$ and $v_j$, $\bm{A}_{ij}=1$, otherwise $\bm{A}_{ij}=0$.

In this research, the anomalous citation detection problem is formulated as a binary classification problem. Specifically, the citation between each paper pair is either a non-anomalous citation or an anomalous one. Matrix $\bm{Y} \in \mathbb{R} ^{|V|\times |V|}$ contains labels for each paper pair, where the value of the element located in its $i$-th row, $j$-th column $y_{ij} = 1$ means that the citation from
paper $i$ to paper $j$ is anomalous, otherwise $y_{ij}$ = 0. TABLE~\ref{Description of notations} provides detailed description of major notations we use in the remaining part.

%$Y \in \mathbb{R} ^{|V|\times |V|}$ contains labels for each paper pair. $y_{ij} = 1$ means that the citation from paper $i$ to paper $j$ is anomalous, otherwise $y_{ij} = 0$.

\begin{table}[ht]
  \centering
  \caption{Description of notations}
  \label{Description of notations}
  \begin{tabular}{ll}
  \toprule
  Notation & \multicolumn{1}{c}{Description} \\
  \midrule
  $G$ &the academic citation network\\
  $V$ &the set of nodes (represent papers)\\
  $E$ & the set of edges (represent citing-cited relationship) \\
  $n_1= |V|$ & the number of nodes in the citation network $G$\\
  $n_2 = |E|$ & the number of edges in the collaboration network $G$\\
  $Y $& the label set for edges\\
  $y_L$ & the labeled citations \\
  $m_{1}$ & the dimension of the node attribute vectors \\
  $m_{2}$ & the dimension of the edge attribute vectors\\
  $d_{1}$ & the dimension of node representation\\
  $d_{2}$ & the dimension of edge representation\\
  $r$ & the learning rate for the autoencoder\\
  $\bm{X}\in \mathbb{R}^{{{n_1} \times m_{1}}}$ & the node local feature matrix\\
  $\bm{Z}\in \mathbb{R}^{{{n_2} \times m_{2}}}$ & relation features for the paper pairs (edge)\\
  $\bm{A}\in \mathbb{R}^{{{n_1} \times n_{1}}}$ & node connection information (adjacency matrix)\\
  $\bm{H^}n \in \mathbb{R}^{{{n_1} \times d_1}}$ & the final node representations \\
  $\bm{H}^e \in \mathbb{R}^{{{n_2} \times d_2}}$ & the final edge representations \\
  \bottomrule
  \end{tabular}
\end{table}

\subsection{Problem Description}
According to the above symbols and definitions, the input of the anomalous citation detection task includes a paper pair $(i,j)$ associated with their local features \bm{$X$}, relation (edge) features \bm{$Z$} for the paper pair, citation context between $i$ and $j$, and partially labeled citations $y_L$. We aim to obtain labels for remaining unmarked citations by the learning and training process. That is, given a paper pair $(i,j)$ and related feature information, GLAD can output the citation relationship between $i$ and $j$.

\section{The GLAD Framework}
\label{sec:Framework}
\begin{figure*}
  \centering
  \includegraphics[width=0.9\textwidth]{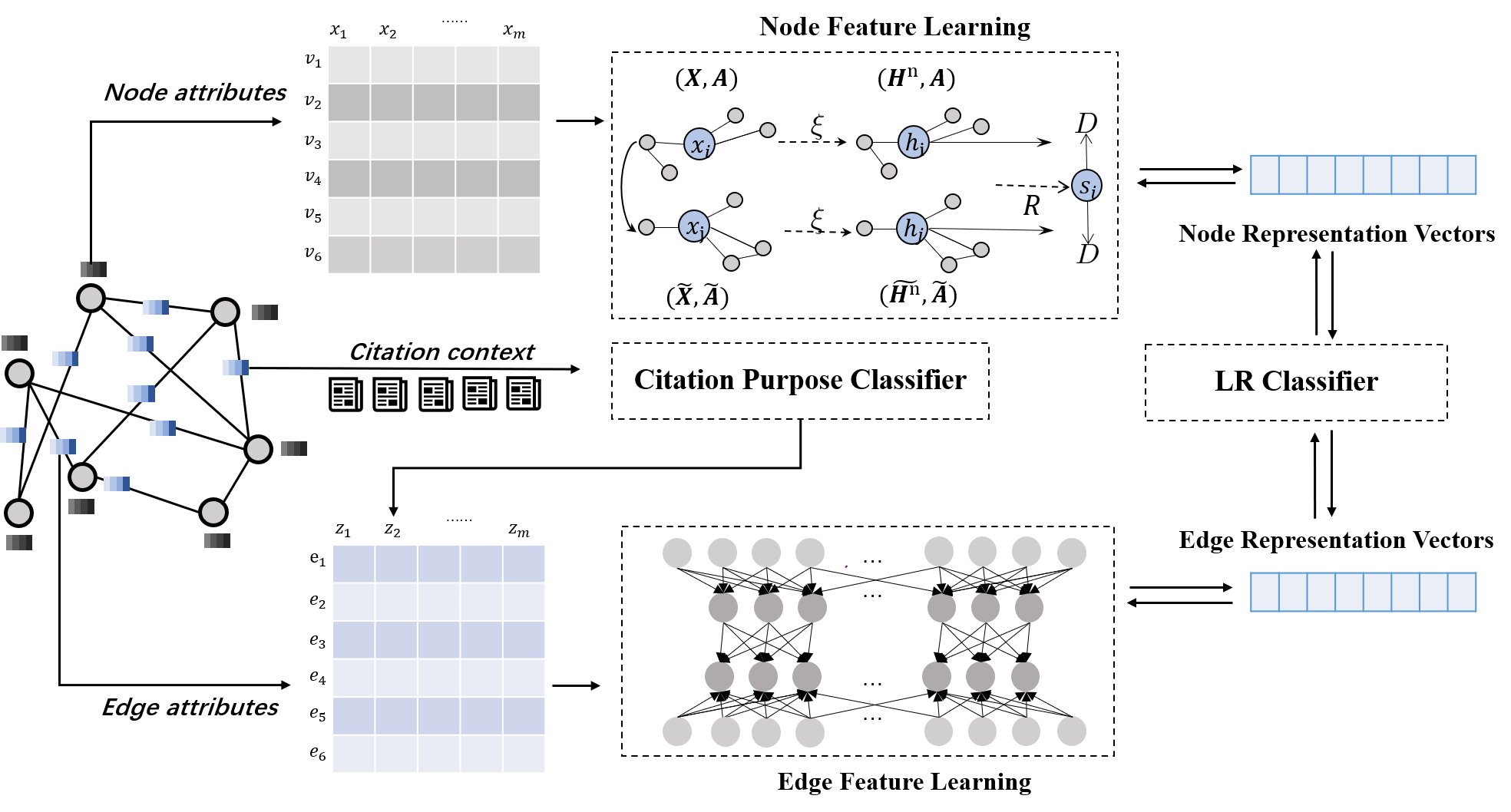}\\
  \caption{The overall framework of GLAD.}
  \label{fig:framework}
\end{figure*}

In this section, we describe the proposed framework for identifying anomalous citations in detail. Specifically, the overall framework contains three main components, i.e., node feature learning, edge feature learning, and anomalous citation identification. Fig.~\ref{fig:framework} gives a graphic example of GLAD. Graph Neural Networks can effectively learn the complex structures of graph-structured data. In the academic citation network, both node and edge attributes can reflect the network properties. For example, from the perspective of node attributes such as the textual information, the research topic and content of a paper are closely related to the abstract of a paper, which helps to judge the similarity of two papers in topic and content. At the same time, edge attributes such as citation purpose and self-citation can reflect the conflict of interest between the two papers. Hence we incorporate attributes for both edges and nodes into GNNs models and formalize the task as modeling the relations between vertices. In order to preserve the global and local network structures in the joint space composed of network structures, node attributes, and edge attributes, we exploit the nodes' and edges' proximity jointly by using the GLAD framework. We hope that the process of feature learning can retain the feature information of both nodes and edges, and at the same time integrate the learned feature representations during the classification process. The individual node and edge feature learning processes are unsupervised. For node feature learning, we select DGI as the workhorse method for our node representation module because it can integrate node attributes and network structure by maximizing mutual information between patch representations and the corresponding graphical high-level summary based on GCN in a completely unsupervised manner. The edge feature learning process is established based on the autoencoder because it can learn the edge feature representations in an unsupervised manner and the and the learning results can be explained by reconstruction. In the following, we elaborate on the specific process of each part and the integrations of these parts.

\subsection{Node Feature Learning}
\label{Node Feature Learning}
Features of nodes can reflect their inherent attributes. In our framework, node features can be regarded as the local properties of papers, such as research topics, and importance in the citation network. These features reflect the importance of papers and topic similarity between paper pairs. For example, cited papers and citing papers should be similar/related in research topics or contents. Node feature learning is carried out from two aspects, textual feature learning and local-global information fusion.

\subsubsection{\textbf{Textual Feature Learning}}
\label{sec:Textual Feature Learning}
Textual information including titles, keywords, abstracts, and main texts is a major component of scientific publications. In order to preserve the local information as well as proximity of each paper from the perspective of textual feature, motivated by Doc2vec~\cite{le2014distributed}, the process of vectorizing text into a numeric space is illustrated as follows.

We first build up a corpus of text information (paper abstracts) as a set $T = \{t_1, t_2,..., t_{|V|}\}$, where $t_i$ is the list of paper $i$'s abstract. In the training process, each abstract is represented by a unique vector in terms of a column of matrix \bm{$D$}$\in \mathbb{R}^{|V|\times p}$, where $|V|$ represents the number of the paragraphs (abstracts) and each abstract is mapped to $p$ dimensions. Each word of the abstract is also represented by a unique vector in terms of a column in matrix \bm{$W$}$\in \mathbb{R}^{N_w\times q}$, where $N_w$ is the number of words in the vocabulary and each word is mapped to $q$ dimensions. The goal is to predict a word with the highest probability in the current context by giving a specific paragraph vector and several word vectors. Specifically, the given sequence of training words will be fixed through a preset window $win$. $win$ represents the maximum distance between the given words and the predicted word in the sentence. The window will traverse the entire corpus. When sliding to position $t$, the abstract vector $d_i$ of the abstract and the word vectors of $|win|$ words in the front and $|win|$ words in the back of the word $w_t$ are used to predict the word $w_t$ that will appear at this position with the maximum probability. The task is to maximize the average log-likelihood:
\begin{equation}
\begin{aligned}
\label{eq1}
&\frac{1}{M-2|win|}\sum^{M-|win|}_{t=|win|}\log p(w_{t-|win|},...,\\
&w_{t-1},w_{t+1},..., w_{t+|win|},d_i)
\end{aligned}
\end{equation}
where $M$ is the number of all training words (the length of the sequence) and $d_i$ is the document representation vector of the abstract list containing the context word in the current window. The prediction task is performed by hierarchical softmax:
\begin{equation}
\begin{aligned}
\label{eq2}
&p(w_t|w_{t-|win|},...,w_{t-1},w_{t+1},..., w_{t+|win|},d_i)\\
&= \frac{\exp(Pr_{w_t})}{\sum_{j=1}^{N_w}\exp(Pr_{w_j})}
\end{aligned}
\end{equation}
where $Pr$ is the log probability of the output, which is computed as:
\begin{equation}
\label{eq3}
Pr=U\bm{a}(w_{t-|win|},...,w_{t-1},w_{t+1},..., w_{t+|win|},d_i;\bm{W}, \bm{D})+b
\end{equation}
where $U$ and $b$ are softmax parameters. $\bm{a}$ is constructed by averaging word vector extracted from \bm{$W$} and document vector extracted from \bm{$D$}. The paper abstracts are vectorized in the latent space by using the PV-DM (Distributed Memory version of Paragraph Vector) model~\cite{le2014distributed}, where similar papers will be close to each other.
%\begin{equation}
%\label{eq0}
%\begin{aligned}
%\frac{1}{2|win|}\sum^{\min(t+|win|,M)}_{\max(t-|win|,1)}
%\log p(w_t|w_{t-|win|},...,\\w_{t-1},w_{t+1},..., w_{t+|win|},d_i)
%\end{aligned}
%\end{equation}

%where $M$ is the number of all training words in the abstract (the length of the sequence). When the window traverses the entire corpus, the average log probability is:

%\begin{equation}
%\begin{aligned}
%\label{eq2}
%&p(w_t|w_{t-|win|},..., w_{t+|win|},d_i)\\
%&= \frac{\exp(w_t^T\cdot(w_{t-|win|},..., w_{t+|win|})+w_t^T\cdot d_i)}{\sum_{j=1}^{T}\exp(w_j^T\cdot(w_{t-|win|},..., w_{t-|win|})+w_j^T\cdot d_i)}.
%\end{aligned}
%\end{equation}

\subsubsection{\textbf{Local-Global Information Fusion}}
Through the textual information learning process illustrated in Section~\ref{sec:Textual Feature Learning}, we have obtained the textual representations for each paper. However, we hope that node representations can reflect not only textual information but also node importance in the citation network. To tackle this problem, inspired by Deep Graph Infomax (DGI)~\cite{velickovic2019deep}, the procedure for the node representations generation is fully summarized as follows.

For nodes in $G$, we first connect their textual representations and artificially defined features illustrated in TABLE~\ref{tab:nodefeature}. Hence, a set of node features $\bm{X}=\{\bm{x}_1, \bm{x}_2,...,\bm{x}_{n_1}\}$ is generated, where $n_1=|V|$. $\bm{A}\in \mathbb{R}^{n_1\times n_1}$ is also provided, where $\bm{A}$ represents node connection information. The final high-level node feature representations are learning through the encoder, $\xi: \mathbb{R}^{n_1\times m_1} \times \mathbb{R}^{n_1\times n_1}\rightarrow \mathbb{R}^{n_1\times d_1}$.
%Hence, the representations can be generated through $\xi(\bm{X},\bm{A})=\bm{X}^{\bm{A}}=\{\bm{x}^{\bm{A}}_{1}, \bm{x}^{\bm{A}}_{2},...,\bm{x}^{\bm{A}}_{n_1}\}$.

\begin{table}[htbp]
\centering
\caption{Description of artificially defined node features}%%%Table caption goes here
\label{tab:nodefeature}
\begin{tabular}{m{1cm}<{\centering} m{7cm}} %%%The number of columns has to be defined here
\hline
Notation & Description \\
\hline
$\bm{d}$& textual representation of the paper\\
\emph{$A_{cic}$ }& citation in-degree centrality for the first author of the paper\\
\emph{$A_{coc}$} & citation out-degree centrality for the first author of the paper\\
\emph{$P_{odc}$} & citation out-degree centrality of the paper\\
\hline
\end{tabular}
\end{table}

The encoder learning process relies on maximizing local mutual information rather than minimizing the construction error. Negative examples are generated by using the corruption function $\chi:(\widetilde{\bm{X}},\widetilde{\bm{A}})= \chi (\bm{X}, \bm{A})$. $\chi$ modifies the network to obtain negative samples. Here, we perturb the node feature matrix by adding row-wise Gaussian noise, while keeping the graph structure unchanged. The patch representations for each node can be learned through graph convolutional networks $\varepsilon$, which can integrate the information of neighbors for the target node. Information integration process can be formulated as:
\begin{equation}
\begin{aligned}
\label{eq4}
\varepsilon(\bm{X},\bm{A})= \sigma(\widehat{\bm{D}}^{-\frac{1}{2}}\widehat{\bm{A}}\widehat{\bm{D}}^{-\frac{1}{2}}\bm{H}_l^n\bm{W}_l^n)
\end{aligned}
\end{equation}
where $\widehat{\bm{A}}=\bm{A}+\bm{I}$. $\widehat{\bm{D}}$ is the degree matrix of $\widehat{\bm{A}}$ and $\bm{H}^n$ is the learning feature of each layer. $\bm{W}^n_l$ is a learnable parameter matrix in the $l$-th layer of node representation learning. For the input layer, $\bm{H}^n_0=\bm{X}$. $\sigma$ is a non-linear activation function. We can get the patch representation $\bm{h}_i^n$ for each node in positive samples set through Eq.(\ref{eq4}). Similarly, we can also generate patch representations $\widetilde{\bm{h}}_i^n$ through Eq.(\ref{eq4}) for negative samples. Global summaries $\bm{s}$ are obtained by passing patch representations through the readout function $\mathcal{R}$:
\begin{equation}
\begin{aligned}
\label{eq5}
\bm{s}=\mathcal{R}(\bm{H}^n)=\frac{1}{N_p}\sum_{i=1}^{N_p}\bm{h}_i^n.
\end{aligned}
\end{equation}
Note that $N_p$ is the number of positive samples. A discriminator $\mathcal{D}$ is employed to discriminate positive and negative patch-summary pairs:
\begin{equation}
\begin{aligned}
\label{eq6}
\mathcal{D}(\bm{h}_i^n,\bm{s})=\sigma({\bm{h}_i^n}^T\bm{W}^n\bm{s}).
\end{aligned}
\end{equation}
where $\bm{W}^n$ is a learnable scoring matrix for the discriminator. The final objective function is:
\begin{equation}
\begin{aligned}
\label{eq7}
L_n=\frac{1}{N_p+N_n}(\sum_{i=1}^{N_p}\mathbb{E}_{(\bm{X},\bm{A})}[\log\mathcal{D}(\bm{h}_i^n,\bm{s})]\\
+\sum_{j=1}^{N_n}\mathbb{E}_{(\bm{\widetilde{X}},\bm{\widetilde{A}})}[1-\log \mathcal{D}(\bm{\widetilde{h}}_j^n,\bm{s})])
\end{aligned}
\end{equation}
where $N_n$ is the number of negative samples. Thus we can generate the representation for each node by integrating local information and global structural information into the final feature representations.

\subsection{Citation Purpose Classification}
\label{citationclassification}
Citation context can provide important semantic information about the relation between the paper and its references, i.e., the authors' intention. The definition of anomalous citations brings the problem of identifying the purpose of the citation. That is, given a target paper $i$ and its citation in a given paper $j$, we need to identify the intention behind selecting $j$ and citing it by the author of $i$.

Previous studies have been devoted to studying different purposes of citing other publications, including comparison, use, criticizing, and so on. Researchers also have come up with different schemes composed of citation purpose categories. In this paper, we adopt a scheme that contains six purpose categories (as shown in TABLE~\ref{Annotation scheme for citation purpose}) after studying the previously used citation taxonomies~\cite{abu2013purpose}.

\begin{table}[htbp]
  \centering
  \caption{Annotation scheme for citation purpose}
    \begin{tabular}{m{2cm}<{\centering} m{5cm}}
    \toprule
    Category & Description \\
    \midrule
    Criticizing  &  The citing paper describes the weakness or strengths of the reference\\
    Comparison & The citing paper compares the present approach or results with the reference\\
    Use  &  The citing paper uses the method, idea or tool of the reference\\
    Substantiating  & Current results, claims of the citing work substantiate or verify the reference or they support each other \\
    Basis &  The reference motivates the work of citing paper \\
    Other  &  The citation doesn't belong to any of the above categories\\
    \bottomrule
    \end{tabular}%
  \label{Annotation scheme for citation purpose}%
\end{table}%

\begin{figure*}
  \centering
  \includegraphics[width=\textwidth]{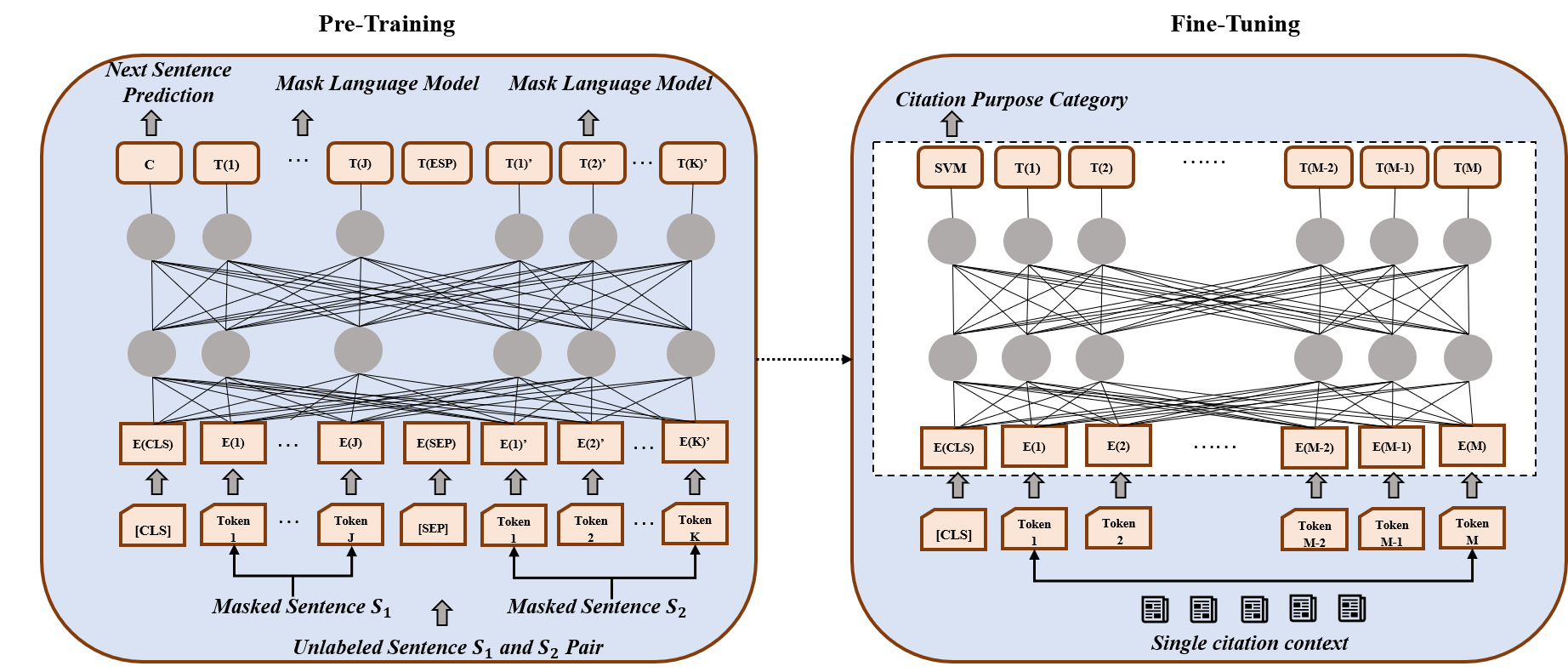}\\
  \caption{The architecture of CPU.}
  \label{fig:cpu}
\end{figure*}

We propose an algorithm to identify the citation purpose based on the citation context, which we refer to as CPU (Citation PUrpose) algorithm. Different from previous work, we incorporate the semantic recognition process into the citation purpose classification task, which is to train the classification model based on feature representations obtained from a set of labeled citation contexts. As shown in Fig.~\ref{fig:cpu}, CPU mainly consists of two parts: feature representation learning process and finetuning process.

To capture semantic information and bidirectional relations in sentences of the citation context, the most important step is to learn a good feature representation for each word. As there are few labeled samples for citation contexts, CPU adopts SCIBERT~\cite{beltagy2019scibert} for pre-training, which is a pre-trained language model established on BERT for scientific text. The core idea of pre-training is to learn good feature representations for words by running a self-supervised learning method on the basis of a massive scientific corpus. Thus we can directly use the feature representations as the word embedding feature of the classification task.

The overall framework for feature representation learning is established on a multi-layer transformer architecture, which is composed of several encoders and decoders. The encoder consists of multi-head attention and a full connection, which is used to convert input corpora into feature vectors. The input of the decoder is the output of the encoder. It is composed of masked multi-head attention, multi-head attention, and a full connection to output the conditional probability of final results. The input of pre-training $\textbf{E}(i)$ is the unit sum of three embedding features, including wordpiece embedding, position embedding, and segment embedding. The architecture for the pre-training process includes two self-supervised tasks:
\begin{enumerate}
 \item Masked Language Model. During training, some words are masked from the input corpus. Then the words are predicted by $\textbf{T}(i)$, the output of the last Transformer at position $i$. In the experiment, 15\% of WordPiece Token~\cite{schuster2012japanese} will be randomly masked. When training the model, a sentence will be input into the model multiple times for parameter learning.
  \item Next Sentence Prediction. This task is to determine whether two sentences follow each other by using the classification token $\textbf{C}$ of the sentence. The training data is generated by randomly extracting two consecutive sentences from the parallel corpus. 50\% of the consecutive sentences are retained as positive samples (IsNext). The second sentence of the remaining 50\% is randomly extracted from the corpus. The relationship between these sentences is ``NotNext".
\end{enumerate}

In finetuning process, we feed final vectors for the token into a linear classification layer. For this purpose, we perform Support Vector Machines (SVM) on citation context representations. Since the problem is a multi-class classification problem, we need to train $N_c$ classifiers, where there are a total of $N_c$ classes of training data. The $i$-th training sample is $(\bm{c}_i,l_i)$, where $\bm{c}_i$ is the textual representation of the citation context learned by CPU. The category label $l_i \in \{1,2,...,N_c\}$. There are $N$ samples. For each category presented in TABLE~\ref{Annotation scheme for citation purpose}, we need to train a binary classifier. Specifically, for each category, we take it as the positive class. All the samples of the remaining $N_c-1$ categories are considered as the negative class. Hence we construct a binary SVM to separate the $j$-th category from the rest of the $N_c-1$ category. It can be regarded as solving a quadratic programming problem:
\begin{equation}
\begin{aligned}
\label{eq8}
    &\min_{\bm{w}^j,b^j,\varepsilon^j}\frac{1}{2}{(\bm{w}^j)}^{T}\bm{w}^j+R\sum^N_{i=1}\varepsilon_i^j\\
    \textrm{s.t.} \quad &(\bm{w}^j)^T\phi(\bm{c}_i)+b^j\geq 1-\varepsilon_i^j, \textrm{if} \ l_i=j\\
    &(\bm{w}^j)^T\phi(\bm{c}_i)+b^j\leq -1+\varepsilon_i^j, \textrm{if} \ l_i\neq j\\
    &\varepsilon_i^j \geq 0
\end{aligned}
\end{equation}
where the subscript $i$ represents the index of the sample. The superscript $j \in \{1,2,...,N_c\}$. $\bm{w}^j$ is a normal vector that determines the direction of the hyperplane. $\phi$ is used to map the input samples to the high-dimensional space. Specifically, we map the sample $\bm{c}_i$ to the feature space and get $\phi(\bm{c}_i)$. $\varepsilon_i^j$ and $R$ represent the slack variable and the regularization coefficient, respectively. The Lagrangian duality of Eq.(\ref{eq8}) is:
\begin{equation}
\begin{aligned}
\label{eq9}
   &\max\sum^N_{i=1}\alpha^j_i-\frac{1}{2}\sum^N_{i,n=1}\alpha^j_i\alpha^j_nl_il_n\langle\phi(\bm{c}_i),\phi(\bm{c}_n)\rangle,\\
    &\textrm{s.t.}\quad 0\leq \alpha^j_i\leq R,i=1,2,...,N;\sum^N_{i=1}\alpha_i^jl_i=0
\end{aligned}
\end{equation}
where $\alpha$ denotes the Lagrangian multiplier of inequality constraints. For $\bm{c}_{new}$, the decision function of type $j$ adopts the following function:
\begin{equation}
\label{eq9-1}
d^j_{new}=(\bm{w}^j)^T\phi(\bm{c}_{new})+b^j=\sum l_i\alpha^j_ik(\bm{c}_i,\bm{c}_{new})+b^j
\end{equation}
where $k$ is the kernel function, $k(\bm{c}_i,\bm{c}_{new})=\phi(\bm{c}_i)^T\phi(\bm{c}_{new})$. For $\bm{c}_{new}$, there are $N_c$ decision functions and $N_c$ outputs. The classification function $f$ is:
\begin{equation}
\label{eq10}
f(\bm{c}_{new})=\arg \max_{j\in \{1,2,...,N_c\} }[\sum l_i\alpha^j_ik(\bm{c}_i,\bm{c}_{new})+b^j]
\end{equation}
By solving the dual problem, the optimal Lagrangian multiplier vector can be obtained. Then we can get the decision function, which is used to classify new data.

\subsection{Edge Feature Learning}
\label{Edge Feature Learning}
Edge attributes can depict relationship intensity. Based on the definition of anomalous citations, edge features are used to help us determine whether there is a clear purpose and excessive relations exist between the cited paper and the citing paper. To preserve the edge proximity, motivated by Tu et al.~\cite{tu2017transnet}, a deep autoencoder is used to obtain feature representations for edges in the network. An autoencoder consists of two parts, i.e., an encoder and a decoder. Thus hidden layers of the autoencoder can continuously encode and decode information obtained from the input data, thereby automatically capturing the characteristics of the input data and keeping them unchanged.

For each edge connecting the citing paper and the cited paper in the citation network, we consider the attributes presented in TABLE~\ref{tab:edgefeature}. Thus we can get the edge feature set represented as $\bm{Z}=\{\bm{z}_1, \bm{z}_2,...,\bm{z}_{n_2}\}$, where $n_2=|E|$.
\begin{table}[htbp]
\centering
\caption{Description of edge features}%%%Table caption goes here
\label{tab:edgefeature}
\begin{tabular}{m{1cm}<{\centering} m{7cm}} %%%The number of columns has to be defined here
\hline
Notation & Description \\
\hline
$\emph{CP}$ & Whether the purpose of the citation is clear or not \\
$\emph{SF}$& Whether the citation is a self-citation or not \\
$\emph{SJ}$ &Whether citing and cited papers are published in the same journal\\
$\emph{JF}$ & The proportion of the journal to which the cited paper published in the reference list of the cited paper \\
$\emph{SI}$ &Whether first authors of citing and cited papers have been working in the same institution\\
$\emph{SR}$&Whether citing and cited papers have same references\\
$\emph{AC}$ &Whether authors of citing and cited papers have collaborated\\
$\emph{CB}$&Whether the first author of the cited paper used to cite citing paper's first author before\\
\hline
\end{tabular}
\end{table}

$\emph{CP}$ is decided by classification results of CPU. If the classification results belong to one of the first five categories (Criticizing, Comparison, Use, Substantiating, and Basis), we treat them as citations with a clear purpose. $\emph{JF}$ represents the proportion of the journal to which the cited paper published in the reference list of the cited paper. It can be calculated as Eq.(\ref{eq11}):
\begin{equation}
\label{eq11}
  \emph{JF} = \frac{|P_{cj}|}{|R_e|}
\end{equation}
where $|P_{cj}|$ is the number of papers in the reference list that publish in the same journal as the cited paper and $|R_e|$ is the number of reference of the citing paper.

The autoencoder adopts $\bm{Z}$ as input. In each hidden layer of the autoencoder, the transformation mechanism can be formulated as:
\begin{equation}
\begin{aligned}
\label{eq12}
   &\bm{h}_{1}^{e} = f (\bm{W}_{1}^e\bm{z} + \bm{b}_{1}^e) \\
   &\bm{h}_{i}^{e} = f(\bm{W}_{i}^e\bm{h}_{i-1}^{e} + \bm{b}_{i}^e), i = 2,...,k
\end{aligned}
\end{equation}
where $f$ is the activation function, i.e., the Sigmoid function. $\bm{W}_{i}^e$ and $\bm{b}_{i}^e$ represent the transformation matrix and the bias vector of the $i$-th layer, respectively. $k$ represents the number of hidden layers. The output of $(i-1)$-th layer is fed into the $i$-th hidden layer. For any arbitrary real value vector, we use the Sigmoid function to map it to range $[0,1]$.

The ultimate goal of edge feature learning is minimizing the reconstruction error between the final feature representation and input. The reconstruction loss is computed as:
\begin{equation}
\begin{aligned}
\label{eq13}
    L_{e} = ||\bm{Z} - \bm{H}^{e}||_{F}^{2}.
\end{aligned}
\end{equation}
By minimizing the distances between the reconstructed representations and the original input, we can get the final edge feature representations.

\subsection{Anomalous Citation Identification}
Now we have obtained node representations and edge representations through the process of feature learning. As mentioned previously, we formulate the task as a binary classification task. Accordingly, we need to train a supervised classifier to predict the label for each paper pair. Specifically, for each unlabeled paper pair $e=(p_i,p_j)$, the final output should be its label (0 or 1). We adopt Logistic Regression (LR) as the classifier. The loss function $L_{lr}$ of LR is:
\begin{equation}
\begin{aligned}
\label{eq14}
    L_{lr} = |L - \sigma(\bm{H} \cdot \bm{W}^{l} + \bm{B}^{l})|
\end{aligned}
\end{equation}
where $\bm{H}$ connects nodes representations and edge representations. $\bm{W}^{l}$ and $\bm{B}^{l}$ refer to the weight matrix and bias matrix, respectively. $\sigma$ is the Sigmoid activation function. For the first-order gradient-based optimization, we use the Adam method.
\subsection{Overall Reconstruction}
To achieve better performance, we jointly train node feature learning, edge feature learning, and classification processes. The joint loss function $L_{sum}$ is computed as:
\begin{equation}
\begin{aligned}
\label{eq15}
    L_{sum} =&  L_{n}  +  L_{e} + \beta L_{lr} + \alpha(\sum_{i}(||\bm{W}_i^e||_F^2+||\bm{B}_i^e||_F^2)\\+&\sum_{l}||\bm{W}_l^n||_F^2)
\end{aligned}
\end{equation}
where $\alpha $ and  $\beta$ are two hyperparameters to adjust the weights of each part. Our goal is to minimize $L_{sum}$. For this purpose, we use back propagation algorithm (BP) with stochastic gradient descent to train the overall framework. The overall architecture of GLAD is summarized in Algorithm 1.
\begin{algorithm}
\label{alg1}
\caption{Anomalous Citation Detection with GLAD}
    \begin{algorithmic}[1] %显示行号,1是每行都显示
    \Require $\bm{A}$, $\bm{X}$, \bm{$Z$}, $\alpha, \beta$, $r$, $L$, and the convergence condition $\epsilon$.
    \Ensure $l$ for each $  e = (i,j)$\\
    \textbf{Initiate}: randomly initiate $\theta_{1} \leftarrow \{\bm{W}^{e}_{i}, \bm{B}^{e}_{i}\}_{i=1}^{m}$,  $\theta_{2} \leftarrow \{\bm{W}^{n}_{l}\}$ , and  $\theta_{3} \leftarrow \{\bm{W}^{l}, \bm{B}^{l}\}$ .\\
    \textbf{Pre-train}: set $r$, conduct edge feature learning process and node feature learning process independently.\\
    \textbf{Train}:
        \While{$\epsilon$ is true}
                \State{randomly generate a batch of data from edge features and local node features}
                \State{calculate $L_{n}$ based on Eq.(\ref{eq7}), get reconstructed $\bm{H}^n$}
                \State{calculate $L_{e}$ based on Eq.(\ref{eq13}), get reconstructed $\bm{H}^e$}
                \State{concatenate $\bm{H} \leftarrow [\bm{H}^n, \bm{H}^e$] }
                \State{calculate $L_{sum}$ based on Eq.(\ref{eq15})}
                \State{update $\theta_{1}, \theta_{2}, \theta_{3}$ }
        \EndWhile
    \end{algorithmic}
\end{algorithm}

\begin{figure*}
  \centering
  \includegraphics[width=0.8\textwidth]{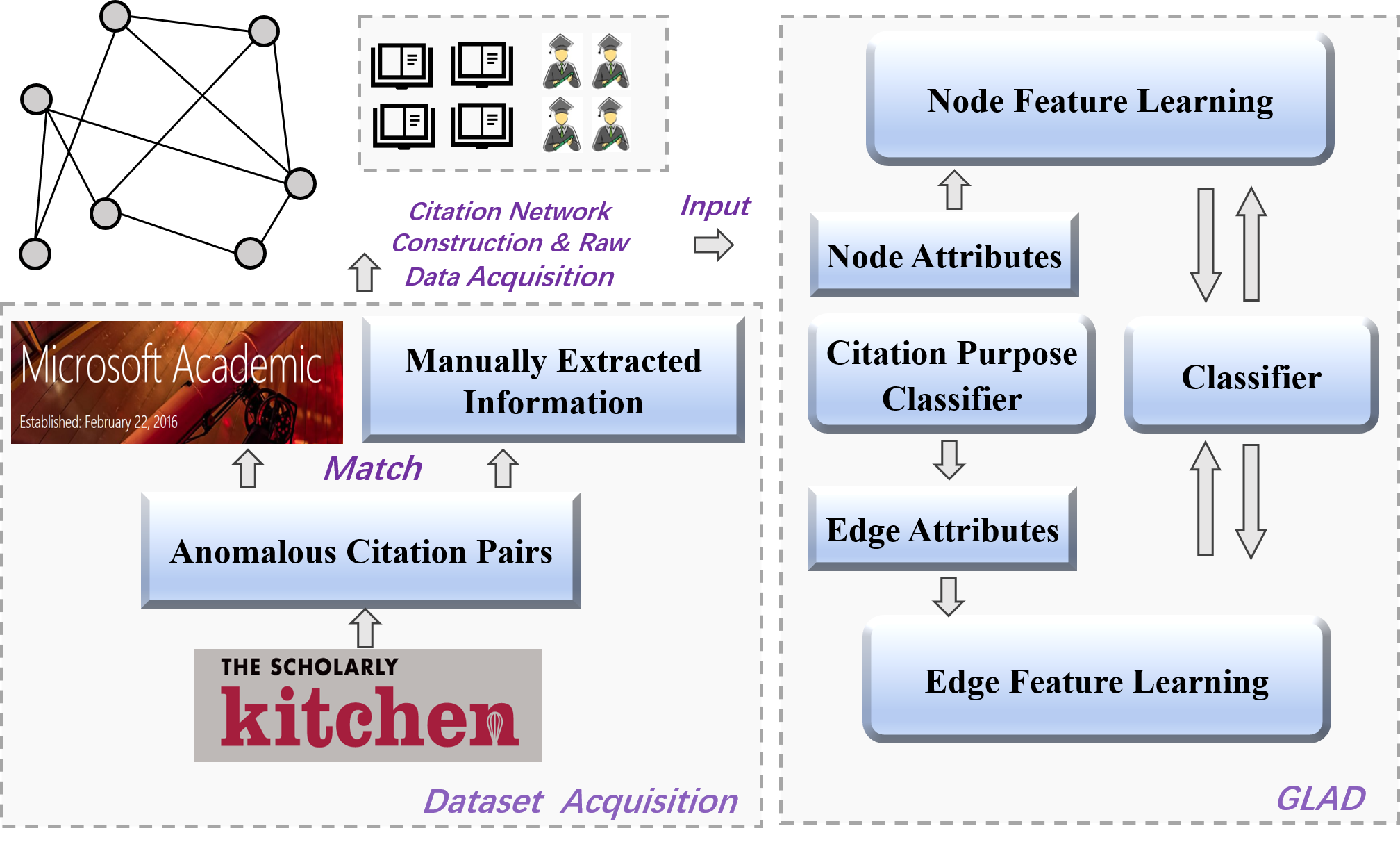}\\
  \caption{Experiment procedure.}
  \label{fig:Experiment procedures}
\end{figure*}

\subsection{Model Complexity Analysis}
The overall framework contains three critical parts, i.e., node feature learning, edge feature learning, and classification. The node representation $\bm{H}^n$ is generating by maximizing mutual information in an unsupervised manner, whose training complexity is $O(c_nn_nI_nd_n (N_p+N_n))$, where $N_p$, $N_n$ represent the number of positive samples and negative samples, respectively. $c_n$ is the sample size (the number of nodes $|V|$). $I_n$ represents the iteration times. $n_n$ is the dimension of node features and $d_n$ is the layer dimension. The edge embedding $\bm{H}^e$ is generated through a deep autoencoder. The complexity of generating the above embedding matrix is $O(c_e n_e d_e I_e)$, where $c_e$ is the size of samples (the number of edges $|E|$), $n_e$ is the dimension of edge features, $d_e$ is the maximum dimension of the hidden layer, and $I_e$ represents the iteration times. The classification process adopts Logistic Regression and the training complexity is $O(n*k)$, where $n$ represents the sample size and $k$ is the feature dimension, which is related to the embedding dimensions. Thus, the overall training complexity of GLAD is $O(c_nn_nI_nd_n (N_p+N_n)+c_e n_e d_e I_e+I_ln*k)$ which is approximate to $O(I_n |V| d_n+I_e |E| d_e+I_l k|E|)$.

\section{Experiments}
\label{sec:Experiments}
In this section, we will introduce the datasets including citation purpose annotation, ground-truth anomalous citations, and the constructed citation network. We validate the effectiveness of GLAD by conducting a series of experiments with discussions of the experimental results. Fig.~\ref{fig:Experiment procedures} presents the overall experiment procedure.

\subsection{Datasets}
\subsubsection{\textbf{Citation Purpose Annotation}}
The annotation process for citation context is conducted by postgraduate students with a good background in natural language processing. We ask them to determine the purpose of citing the target reference by choosing from the six purpose categories described in TABLE~\ref{Annotation scheme for citation purpose}. To estimate inter-annotator agreement, each citation context is assigned to three different annotators. Kappa Coefficient~\cite{cohen1968weighted} is used to measure the agreement:
\begin{equation}
\label{eq1}
K = \frac{P(A)-P(E)}{1-P(E)}
\end{equation}
where $P(A)$ represents the relative observed agreement among three annotators and $
P(E)$ represents the hypothetical probability of agreement.

In our task, the agreement among three annotators are $K_{12} =0.606$, $K_{13} =0.477$, and $K_{23} =0.677$, respectively, which indicates moderate and substantial agreements. Besides, uncertain labeled results are discarded. The distribution of the purpose categories are shown in TABLE~\ref{citation purpose}.

\begin{table}[htbp]
  \centering
  \caption{The proportion of citation purpose in annotation}
    \begin{tabular}{m{2.3cm} m{2.3cm}<{\centering} m{2.3cm}<{\centering}}
    \toprule
    Categories & Number & Proportion\\
    \midrule
    Criticizing &150 & 14\%\\
    Comparison & 98&9\%\\
    Use & 183&17\%\\
    Substantiating &43& 4\%\\
    Basis & 323& 30\%\\
    Neutral/ Other &280& 26\% \\
    \bottomrule
    \end{tabular}%
  \label{citation purpose}%
\end{table}%

\subsubsection{\textbf{Anomalous Citation Dataset}}
For anomalous citation detection in academic networks, ground-truth anomalous citations are not clearly defined. Hence, we artificially create a dataset motivated by previous research~\cite{kojaku2020detecting}. The dataset generation process is described as follows in detail.

We first collect the paper titles which are reported as papers causing ``journal cartels" in~\cite{davis2012emergence}. Most of these papers are published in donor journals including the \emph{Medical Science Monitor}, and \emph{The Scientific World Journal}. These papers cite papers published in \emph{Cell Transplantation} at an excessively high rate. For example, the paper titled ``Regenerative medicine for neurological disorders" published in \emph{The Scientific World Journal} has cited 124 papers in total, 96 of which are published in \emph{Cell Transplantation}. Two of the four authors of that paper are editors of \emph{Cell Transplantation}. The impact factor of \emph{Cell Transplantation} has dropped by removing the reference of that paper. We also crawl the citation cartels detection results from previous research~\cite{kojaku2020detecting}. Taking the paper ``Crystal structure refinement with SHELXL?" published in \emph{Acta Crystallographica C} as an example. The paper has received 594 citations from the donor journals \emph{IUCrData} and \emph{Acta Crystallographica Section E}, accounting for 94\% of the total citations in 2017. Removing the citations from donor journals decreases the JIF of \emph{Acta Crystallographica C} by 22\%. Details of the anomalous citations dataset are described in TABLE~\ref{anomaly}.

\begin{table*}[htbp]
  \centering
  \caption{Details of anomalous datasets}
    \begin{tabular}{m{5cm}<{\centering} m{3.5cm}<{\centering} m{4.5cm}<{\centering} m{1.5cm}<{\centering} m{1.5cm}<{\centering}}
    \toprule
    Paper Title & Publishing  Journal & Anomalous Citation Relationship & Year  & State \\
    \midrule
    Stem cells have the potential to rejuvenate regenerative medicine research & \emph{Medical Science Monitor }& References to papers published in \emph{Cell Transplantation} and \emph{Medical Science} & 2008-2009 & Retracted \\
    Regenerative medicine for neurological disorders & \emph{The Scientific World Journal} & References to papers published in \emph{Cell Transplantation} and \emph{The Scientific World Journal}  & 2008-2009 &Retracted  \\
    The continued promise of stem cell therapy in regenerative medicine & \emph{Medical Science Monitor} & References to papers published in \emph{Cell Transplantation}  and \emph{Medical Science Monitor} & 2009-2010 & Retracted \\
    A showcase of bench-to-bedside regenerative medicine at the 2010 ASNTR & \emph{The Scientific World Journal} & References to papers published in \emph{Cell Transplantation}  and \emph{The Scientific World Journal}  & 2010-2011 & Retracted \\
    Technology and innovation: 2010 a year in review & \emph{Cell Transplantation} & References to papers published in \emph{Technology and Innovation} and \emph{Cell Transplantation} & 2009-2010 & Retracted \\
    Crystal structure refinement with SHELXL? & \emph{Acta Crystallographica Section C} & Citations from \emph{IUCrData} and \emph{Acta Crystallographica Section E} & 2017  & Normal \\
    The SCARE 2018 statement: Updating consensus Surgical CAse REport (SCARE) guidelines &  \emph{International Journal of Surgery} & Citations from \emph{Annals of Medicine and Surgery} and \emph{International Journal of Surgery Case Reports} & 2019  & Normal \\
    Regulatory Intermediaries in the age of Governance & \emph{Annals of American Academy of Political and Social Science } & Citations from Annals of\emph{ Medicine and Surgery} and \emph{International Journal of Regulation and Governance} & 2019  & Normal \\
    \bottomrule
    \end{tabular}%
  \label{anomaly}%
\end{table*}%

\subsubsection{\textbf{Microsoft Academic Search Dataset}}
The citation relationships of anomalous papers are obtained from a widely used bibliographic dataset, Microsoft Academic Graph (MAG)\footnote{https://www.openacademic.ai/oag/}. It contains abundant information for each paper such as journal name, publication year, authors, abstract, citation relationship, and so on. In particular, it provides citation context, that is, the raw text where the citation is mentioned in the paper. We use citation contexts to conduct citation purpose classification process as mentioned in Section~\ref{citationclassification}. We match the anomalous papers mentioned in TABLE~\ref{anomaly} and construct the citation network centered on them. Edges for anomalous citations are labeled as anomalous, and the rest are normal. The features of papers (such as abstract) and citation relationships (such as citation context) can also be obtained and calculated in this process. It's important to note that the dataset doesn't contain some retracted papers. We searched for the corresponding papers on the Internet and then added them back manually. Finally, the constructed network contains 4718 citing-cited relationships, of which 300 citations are anomalous.
%TABLE~\ref{The statistics of datasets} lists the statistics of the constructed citation network.
%\begin{table}[htbp]
%  \centering
%  \caption{The statistics of the constructed citation network}
%    \begin{tabular}{m{3cm} m{3cm}<{\centering} }
%    \toprule
%    Item & Number \\
%    \midrule
%    Publications & 5158   \\
%    Citing-cited relationships & 5296  \\
%    Anomalous citations  & 1269   \\
%    Normal citations & 3027\\
%    \bottomrule
%    \end{tabular}%
%  \label{The statistics of datasets}%
%\end{table}%

\subsection{Evaluation Metrics}
\label{baseline}
In this paper, the task of anomalous citation identification is formulated as a binary classification problem, i.e., for each paper pair $(p_i,p_j)$, whether $(p_i,p_j)$ is anomalous (yes or no). Therefore, we adopt four widely used metrics for classification tasks, including Accuracy, Precision, Recall, and F1-score to validate the performance of the model. Specifically, before the task of anomalous citation identification, we divide the training set into two categories:
\begin{itemize}
  \item Positive sample: The positive sample set $P$ contains all real anomalous paper pairs (labeled as 1);
  \item Negative sample: The negative sample set $N$ contains all sample pairs who are not anomalous paper pairs (labeled as 0).
\end{itemize}
After the classification process, we assume that set $TP$ includes all pairs in $P$ that are correctly classified as anomalous citations, and $FN$ are misclassified pairs in $P$. In the same way, the set $FP$ involves the misjudged pairs in $N$ and $TN$ is the correctly judged set of $N$. Apparently, $N = FP + TN$. Hence, the indicators can be calculated as follows:
\begin{itemize}
  \item $Accuracy = \displaystyle\frac{|TP| + |TN|}{|P|+|N|}$
  \item $Precision = \displaystyle\frac{|TP|}{|TP|+|FP|}$
  \item $ Recall = \displaystyle\frac{|TP|}{|TP|+|FN|}$
  \item $F1-score= \displaystyle\frac{2*Precision*Recall}{Precision+Recall}$.
\end{itemize}
Furthermore, we also utilize ROC-AUC (AUC) to evaluate the overall performance of all methods. AUC is calculated based on the relative ranking of the predicted probabilities of all instances.

\begin{table*}[htbp]
  \centering
  \caption{Performance comparison of different anomaly detection methods based on graph learning on the task of anomalous citation detection}
%    \resizebox{0.5\textwidth}{!}{
    \begin{tabular}{l|l|l|l|l|l}
    \toprule
    \diagbox{Methods}{Metrics} & AUC-ROC & Accuracy & Precision & Recall & F1-score \\
    \midrule
    GDN   & 0.7431$\pm$0.015 & 0.6254$\pm$0.0054 & 0.5731$\pm$0.0038 & 0.7315$\pm$0.0043 & 0.6313$\pm$0.0037 \\
    \midrule
    Meta-GDN & 0.7653$\pm$0.013 & 0.6233$\pm$0.0022 & 0.6152$\pm$0.0025 & 0.7506$\pm$0.0112 & 0.6579$\pm$0.0046 \\
    \midrule
    OC-GCN & 0.6531$\pm$0.011 & 0.5437$\pm$0.0036 & 0.5229$\pm$0.0078 & 0.8237$\pm$0.0057 & 0.6312$\pm$0.0044 \\
    \midrule
    OC-GAT & 0.8115$\pm$0.012 & 0.8431$\pm$0.0012 & 0.8345$\pm$0.0023 & 0.8441$\pm$0.0011 & 0.8564$\pm$0.0034 \\
    \midrule
    OC-SAGE & 0.8652$\pm$0.012 & 0.8646$\pm$0.0065 & 0.8843$\pm$0.0037 & 0.8542$\pm$0.0036 & 0.8465$\pm$0.0045 \\
    \midrule
    CARE-Att & 0.8935$\pm$0.022 & 0.8754$\pm$0.0012 & 0.9012$\pm$0.0122 & 0.8534$\pm$0.0014 & 0.8653$\pm$0.0043 \\
    \midrule
    CARE-Weight & 0.8963$\pm$0.014 & 0.8644$\pm$-0.0025 & \textbf{0.8754}$\pm$0.0017 & 0.8743$\pm$0.0034 & 0.8853$\pm$0.0055 \\
    \midrule
    CARE-Mean & 0.8812$\pm$0.012 & 0.8623$\pm$0.0067 & 0.8671$\pm$0.0056 & 0.8586$\pm$0.0049 & 0.8785$\pm$0.0027 \\
    \midrule
    CARE-GNN & \textbf{0.8983}$\pm$0.012 & 0.8825$\pm$0.0034 & 0.8945$\pm$0.0015 & 0.8432$\pm$0.0026 & 0.8654$\pm$0.0033 \\
    \midrule
    GLAD  & 0.8953$\pm$0.0014 & \textbf{0.913}$\pm$0.0087 & \textbf{0.9683}$\pm$0.0054 & 0.854$\pm$0.0109 & \textbf{0.9075}$\pm$0.0090 \\
    \bottomrule
    \end{tabular}%
  \label{baselines2}%
\end{table*}%

\subsection{Baselines}
\label{baseline}
Due to the lack of any existing baseline for the given task, we compare GLAD with the following state-of-the-art models of graph neural networks because our feature learning process is built on graph neural networks.
\begin{itemize}
  \item \textbf{DeepWalk~\cite{perozzi2014deepwalk}}. Deepwalk uses the co-occurrence relationship between nodes in the network to learn node representations. The latent representations are generated by employing skip-gram.
  \item \textbf{Node2vec~\cite{grover2016node2vec}}. Node2vec obtains node sequences through random walk in terms of both DFS and BFS. Hence the low-dimensional feature representations can comprehensively preserve the similarity of DFS and BFS neighborhoods.
  \item \textbf{LINE~\cite{tang2015line}}. LINE is also a graph embedding method based on the assumption of neighborhood similarity. LINE uses BFS to construct neighborhoods. In addition, LINE can also be applied in weighted graphs.
  \item \textbf{GraphSAGE~\cite{hamilton2017inductive}.} GraphSAGE is an inductive framework, which can use vertex feature information (such as text attributes) to efficiently generate embedding for nodes.
  \item \textbf{DGI~\cite{velickovic2019deep}.} DGI learns node representations by maximizing mutual information between the graph-level summary representation and the local patches. DGI is an unsupervised network representation learning framework.
  \item \textbf{MLP}. MLP (Multi-layer perceptron) neural network is a common ANN (artificial neural network) algorithm, which consists of an input layer, an output layer, and one or more hidden layers. Different layers of MLP neural network are fully connected.
\end{itemize}

The comparison results are presented in Fig.~\ref{baselines}. We also compare GLAD with state-of-the art methods for anomaly detection based on graph learning:
\begin{itemize}
  \item \textbf{GDN~\cite{ding2021few}}. GDN is a new GNN architecture designed for network anomaly detection with limited labeled data. Specifically, the authors also propose Meta-GDN, which integrates cross-network meta-learning to detect anomalies with few labeled instances.
  \item \textbf{OCGNN~\cite{wang2021one}}. OCGNN is a one-class classification framework that contains a series of algorithms for graph anomaly detection. OCGNN combines the powerful representation ability of GNNs along with the classical one-class objective. In this work, we compare GLAD with OC-GCN, OC-GAT, and OC-SAGE.
  \item \textbf{CARE~\cite{dou2020enhancing}}. CARE enhances the GNN aggregation process with three unique modules against camouflages including a label-aware similarity measure, a similarity-aware neighbor selector, and a relation-aware neighbor aggregator. In this work, we compare GLAD with various forms of CARE, including CARE-Att, CARE-Weight, CARE-Mean, and CARE-GNN.
\end{itemize}

Table~\ref{baselines2} lists the results of comparing GLAD with baselines. For the sake of fair comparisons with GLAD which considers both node attributes and edge attributes, we concatenate the raw edge attribute matrix $\bm{Z}$ to learned node embeddings of the methods that ignore the edge attributes such as Deepwalk, LINE, Node2vec, and DGI, i.e., $\bm{H}\leftarrow[\bm{H}^n;\bm{Z}]$. Furthermore, we also compare GLAD with different variants to examine the learning efficacy of each module and eliminate the effect of each module. TABLE~\ref{Summary of the detection methods for comparison} presents the details of each variant. First, we only utilize node feature learning process mentioned in Section~\ref{Node Feature Learning} (GLAD-N) and edge feature learning mentioned in Section~\ref{Edge Feature Learning} (GLAD-E) to detect anomalous citations. To examine whether node and edge features are useful, we use random vectors to replace the node features in the process of node feature learning (GLAD-RN) and the edge features in the process of edge feature learning (GLAD-RE). We then remove the loss function of training nodes features $L_n$ (GLAD-EL) and the loss function of training edge features $L_e$ in the final loss function (GLAD-NL) to observe the model performance without auxiliary training objective. Finally, we also use the edge feature learning module (based on the autoendoder) to learn the node features and connect the embeddings with edge representations as the final input (GLAD-EE).
\begin{table*}[htbp]
  \centering
  \caption{Summary of variants for comparison}
  \label{Summary of the detection methods for comparison}
  \begin{tabular}{m{1.8cm}<{\raggedright} m{1.8cm}<{\centering} m{1.8cm}<{\centering}m{1.8cm}<{\centering} m{1.8cm}<{\centering} m{1.8cm}<{\centering}m{1.8cm}<{\centering} m{2cm}<{\centering}}
  \toprule
  Variants of GLAD & Node Feature Learning& Edge Feature Learning& Random Vectors for Node Features &Random Vectors for Edge Features& Loss Function for Node Learning & Loss Function for Edge Learning&Edge Learning Module for Node Representation\\
  \midrule
  GLAD-N   &$\surd$ &~ &~&~&$\surd$ &$\surd$ &~             \\
  GLAD-E & ~          &$\surd$ &~ &~&$\surd$&$\surd$  &~       \\
  GLAD-RN& $\surd$          &$\surd$ &$\surd$ &~&$\surd$&$\surd$ &~\\
  GLAD-RE& $\surd$          &$\surd$ &~ &$\surd$&$\surd$&$\surd$&~\\
  GLAD-EL& $\surd$          &$\surd$ &~ &~&~&$\surd$&~\\
  GLAD-NL& $\surd$          &$\surd$ &~ &~&$\surd$&~&~\\
  GLAD-EE& $\surd$          &$\surd$ &~ &~&$\surd$&$\surd$&$\surd$\\
  GLAD   &$\surd$           &$\surd$ &~ &~&$\surd$&$\surd$&~         \\
  \bottomrule
  \end{tabular}
\end{table*}

\begin{table*}[htbp]
  \centering
  \caption{Citation purpose classification results of CPU using different classifiers }
    \begin{tabular}{m{2.5cm}<{\centering} |m{3cm}<{\centering} |m{1.5cm}<{\centering} |m{1.5cm}<{\centering} |m{1.5cm}<{\centering} |m{1.5cm}<{\centering} |m{1.5cm}<{\centering} |m{1.5cm}<{\centering} }
    \toprule
    Methods &  \diagbox{Metrics}{Categories} & Criticizing & Comparison & Use & Substantiating & Basis & Neutral/Other \\
    \midrule
    \multirow{3}[2]{*}{SVM} & Precision & 64.30\% & 58.10\% & 57.90\% & 59.90\% & 45.70\% & 54.60\% \\
          & Recall & 79.20\% & 51.70\% & 67.30\% & 64.10\% & 44.60\% & 77.30\% \\
          & F1-score & 70.70\% & 56.30\% & 66.40\% & 60.90\% & 44.90\% & 67.60\% \\
    \midrule
    \multirow{3}[2]{*}{Logistic Regression} & Precision & 61.10\% & 49.90\% & 66.80\% & 52.10\% & 46.70\% & 52.30\% \\
          & Recall & 68.40\% & 55.60\% & 52.50\% & 43.50\% & 44.10\% & 65.40\% \\
          & F1-score & 63.30\% & 51.90\% & 55.80\% & 47.50\% & 44.30\% & 55.10\% \\
    \midrule
    \multirow{3}[2]{*}{Naive Bayes} & Precision & 67.70\% & 59.90\% & 33.20\% & 50.60\% & 55.30\% & 49.90\% \\
          & Recall & 69.00\% & 56.40\% & 45.90\% & 66.70\% & 42.10\% & 54.20\% \\
          & F1-score & 68.10\% & 57.00\% & 39.60\% & 59.40\% & 48.30\% & 52.60\% \\
    \bottomrule
    \end{tabular}%
  \label{tab:cpuresults}%
\end{table*}%

\begin{table*}[htbp]
  \centering
  \caption{The performance of GLAD with different batch size}
    \begin{tabular}{m{3cm}<{\centering} |m{2cm}<{\centering}| m{2cm}<{\centering} |m{2cm}<{\centering} |m{2cm}<{\centering} |m{2cm}<{\centering} |m{2cm}<{\centering} }
    \toprule
    \diagbox{Metrics}{ Batch Size}   & 1 & 4& 8 & 16&32&64\\
    \midrule
    Accuracy & 0.8958 $\pm$ 0.009 & \textbf{0.8965} $\pm$ 0.0098& 0.8943 $\pm$ 0.0099 & 0.8953 $\pm$ 0.0098 & 0.8958 $\pm$ 0.0098 & 0.8954 $\pm$ 0.0098 \\
    \midrule
    Precision & 0.9547 $\pm$ 0.006  & \textbf{0.9552} $\pm$ 0.0066 & 0.9522 $\pm$ 0.0068 & 0.9542 $\pm$ 0.0067 &\textbf{0.9552} $\pm$ 0.0066 &0.9545 $\pm$ 0.0067\\
    \midrule
    Recall &0.8309 $\pm$ 0.012 & \textbf{0.8884} $\pm$ 0.010 &0.8871 $\pm$ 0.0102 & 0.8305 $\pm$ 0.0121 & 0.8305 $\pm$ 0.0121 & 0.8305 $\pm$ 0.0121 \\
    \midrule
    F1-score   & 0.8885 $\pm$ 0.0101 & \textbf{0.8957} $\pm$ 0.0098 & 0.8943 $\pm$ 0.0099 & 0.8881 $\pm$ 0.010 & 0.8885 $\pm$ 0.0101& 0.8882 $\pm$ 0.0101\\
    \bottomrule
    \end{tabular}%
  \label{tab:batchsize}%
\end{table*}%

\subsection{Results and Analysis}
The experimental results are presented together with a number of case studies from the following three perspectives, including (1) comparison against baselines, (2) comparison of GLAD variants, and (3) parameter sensitivity.

\subsubsection{\textbf{Comparison against Baselines}}

\begin{figure}
  \centering
  \includegraphics[width=0.5\textwidth]{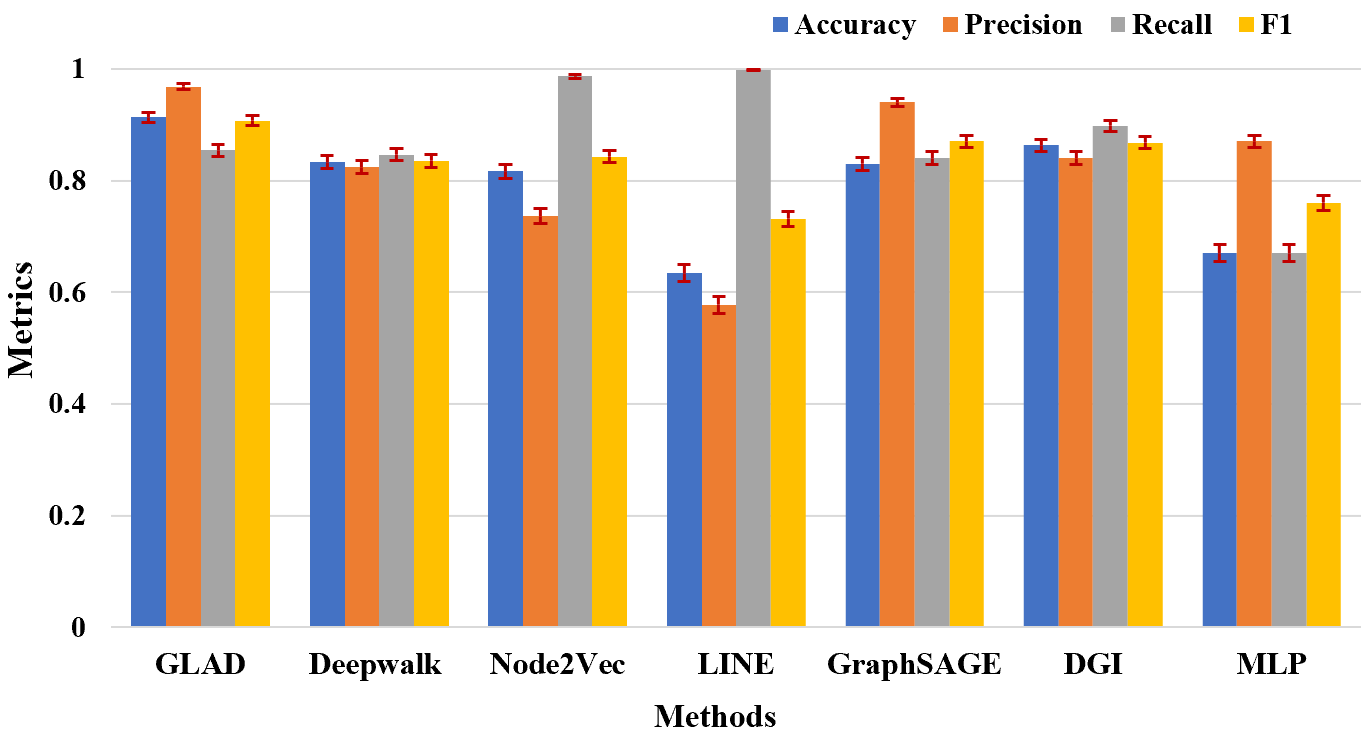}\\
  \caption{Performance comparison of different methods on the task of anomalous citation detection.}
  \label{baselines}
\end{figure}

The results of comparing GLAD with all baselines mentioned in Section~\ref{baseline} are shown in Fig.~\ref{baselines}. From the results, we can observe that Our method GLAD outperforms all the baselines in terms of Accuracy, Precision, and F1-score, which shows the effectiveness of GLAD. Although the results of the baselines on Recall are higher than the proposed method, their Precisions are lower. Higher Precisions indicate that when we use the model to determine that a citation is an anomalous citation, then it is highly likely to be anomalous. We don't expect to find all anomalous citations but hope that the identified anomalous citations are as accurate as possible. In addition, GLAD has achieved the best results on the comprehensive indicator F1-score. In all the comparison methods, DGI and GraphSAGE have better performances, which illustrates that learning attribute features is more effective in the task of anomalous citation detection. The comparison between graph neural networks and MLP shows that feature learning based on neural networks can improve the performance of the model. Compared with directly inputting features, learning features first can effectively improve the performance of anomalous citation detection.

The results of comparing GLAD with all baselines for anomaly detection mentioned in Section~\ref{baseline} are presented in TABLE~\ref{baselines2}. Among all GNN baselines in TABLE~\ref{baselines2}, in terms of Accuracy, Precision, and F1-score, our approach outperforms all the other compared methods by a significant margin. It demonstrates that GLAD can better detect abnormal edges than other methods. GLAD is capable of extracting comprehensive knowledge across the citation network, which further enhances the detection performance on the target network. Besides, experimental results have shown that the proposed model performs better than OC-based model. One-class classification based methods attempt to model normal examples and classify new examples as either normal or abnormal. In this paper, we aim to detect anomalous citations through analyzing the contexts, and give an explicit definition of anomalous citation shown in Section~\ref{sec:Problem Definition}. In real-world scenarios, there are different kinds of citation purposes between two papers. In other words, it is impossible to learn all features of normal citations due to the complexity and diversity of normal samples. Therefore, the one-class classifier is not suitable for detecting anomalous citations.

\subsubsection{\textbf{Comparison of GLAD Variants}}
\begin{figure}[htbp]
  \centering
  \includegraphics[width=0.5\textwidth]{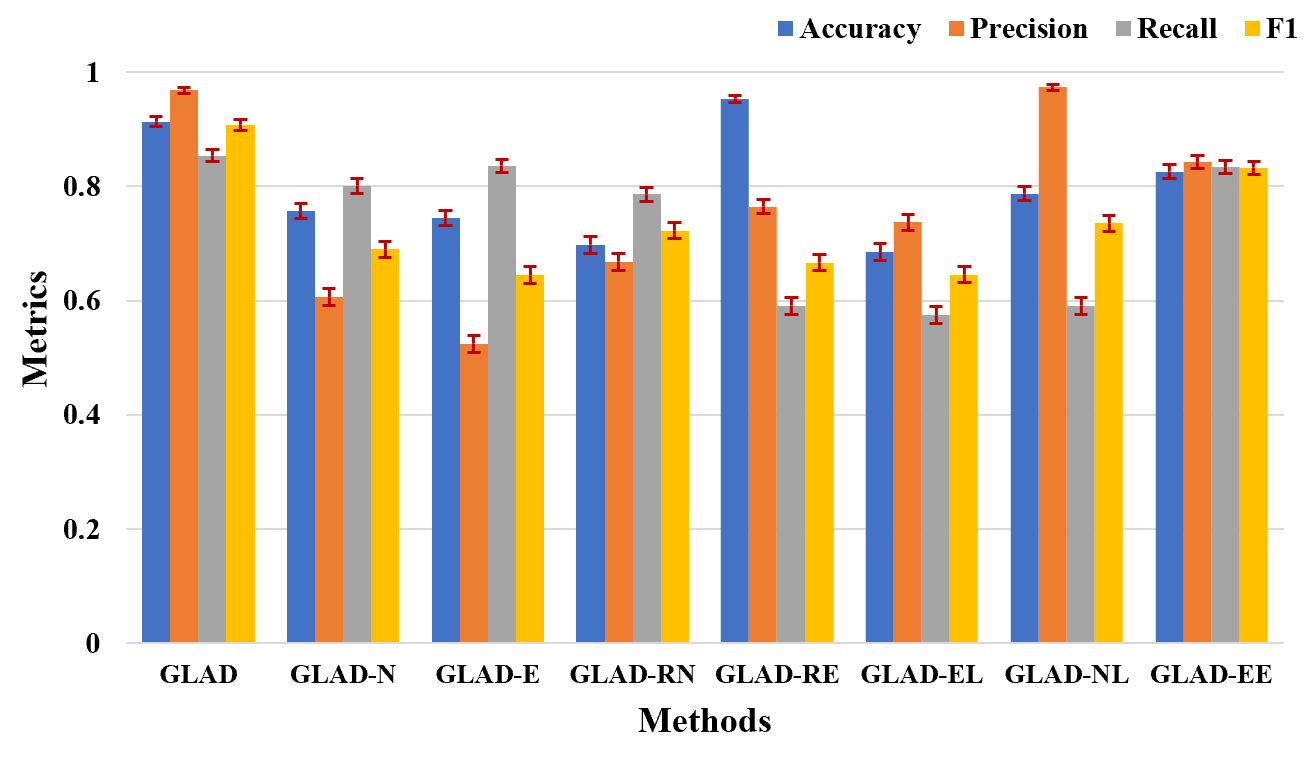}\\
  \caption{Performance comparison of different variants on the task of anomalous citation detection.}
  \label{variants}
\end{figure}

A key ingredient of GLAD is the feature learning of both node attributes and edge attributes. To validate the effectiveness of this mechanism, here we examine the performance of GLAD with only node feature learning process or with only edge feature learning process. From Fig.~\ref{variants}, From the figure, we can clearly observe that GLAD is 36.2\% and 44.4\% higher than GLAD-N and GLAD-E respectively on Precision. GLAD is 21.7\% and 26.3\% higher than GLAD-N and GLAD-E in terms of F1-score, respectively. In general, the performance of GLAD is better than GLAD-N and GLAD-E, which suggests that utilizing both node and edge attributes in feature learning processes is necessary, and GLAD can facilitate them well to help detect anomalous citations. Besides, if we use random vectors for feature learning (GLAD-RN and GLAD-RE), the performances will be better than that of GLAD-N and GLAD-E on F1-score, but there is still a certain gap compared with GLAD. The proposed features are more effective than random vectors. If we do not minimize the loss function of feature learning (but still use these features), the performance of the models will be further improved, but will not exceed GLAD. Without the auxiliary training objectives, the model performs worse. Finally, using the autoencoder to learn the node features and edge features at the same time (GLAD-EE), the best results are obtained in all variants. This shows the necessity of feature learning for both edges and nodes. However, in the task of anomalous citation detection, the performance of node learning with autoencoder is not as good as that of DGI. In conclusion,  although some indicators of some variants exceed GLAD, the overall performance of GLAD is better than that of all variants, which indicates that the proposed features are effective and it is necessary to conduct feature learning processes.

\begin{table*}[htbp]
  \centering
  \caption{The performance of GLAD with different embedding dimension}
    \begin{tabular}{m{6cm}<{\centering} |m{1.8cm}<{\centering}| m{1.8cm}<{\centering} |m{1.8cm}<{\centering} |m{1.8cm}<{\centering} |m{1.8cm}<{\centering}}
    \toprule
    \diagbox{Metrics}{Embedding Dimension}   & 32 & 64& 128& 256&512\\
    \midrule
  Accuracy & 0.8913 $\pm$ 0.0150 & 0.8947 $\pm$ 0.0148 & 0.8967 $\pm$ 0.0146 & 0.8965 $\pm$ 0.0146 &\textbf{ 0.897} $\pm$ 0.0146 \\
      \midrule
   Precision & 0.96 $\pm$ 0.0095 & \textbf{0.9641} $\pm$ 0.0090 & 0.96 $\pm$ 0.0094& 0.9627 $\pm$ 0.0091 & 0.9622 $\pm$ 0.0091 \\
       \midrule
   Recall & 0.8165 $\pm$ 0.0187 & 0.82 $\pm$ 0.0185 & \textbf{0.828} $\pm$ 0.0181 & 0.825 $\pm$ 0.0183 & 0.8265 $\pm$ 0.0182 \\
       \midrule
   F1-score    & 0.8825 $\pm$ 0.0156 & 0.8862 $\pm$ 0.0153 & 0.8891 $\pm$ 0.0151 & 0.8885 $\pm$ 0.0151 & \textbf{0.8892} $\pm$ 0.0151\\
   \bottomrule
   \end{tabular}%
  \label{tab:embedding dimension}%
\end{table*}%

\subsubsection{\textbf{Parameter Sensitivity}}

We conduct a series of experiments to examine how different parameters influence the performance of GLAD. We report the performance of GLAD with respect to the following parameter settings: 1) batch size; 2) embedding dimension; 3) learning rate; 4) number of hidden layers and units in the autoencoder.
For each parameter setting, we fix other parameters and try 10 different splits of the training and testing. The results are reported as the average performance of the 10 experiments. The results also include confidence intervals of each index.

% Table generated by Excel2LaTeX from sheet 'Sheet7'
First of all, TABLE~\ref{tab:cpuresults} presents the citation purpose classification results of CPU using different classifiers including SVM, Logistic Regression, and Naive Bayes. By comparing the results of experiments, we can find that the SVM performs better in the task of citation purposes classification. Besides, one of the edge features $\emph{CP}$ refers to that whether the purpose of the citation is clear or not. In this paper, we also try binary classification. The classification precision can be improved to 78.4\%.

\textbf{Batch Size.} TABLE~\ref{tab:batchsize} lists the experimental results of GLAD's performance when hyperparameter batch size is set doubled increasing from 1 to 64. We can observe that the performance of GLAD is relatively stable within a large range of batch sizes. In addition, the performance of GLAD achieves the best performance in Accuracy, Precision, F1-score, and Recall when the batch size is set to 4. This observation demonstrates that larger batch sizes don’t necessarily lead to better performance.

\textbf{Embedding Dimension.} We empirically evaluate the impact of edge embedding dimension. TABLE~\ref{tab:embedding dimension} illustrates the performance of GLAD when altering the edge embedding dimension from 32 to 512. We can conclude that the performance of GLAD slightly fluctuates within a large range of edge embedding dimensions, and the performance drops when the edge embedding dimension is too small. The model can reach a relatively better state when the embedding dimension is set to 512.

\textbf{Learning Rate.} Learning rate controls the update speed of GLAD. TABLE~\ref{tab:learning rate} shows how different values of the learning rate influence the performance of GLAD. The results show that the performance drop when the learning rate is either too large or too small. We can see that the proposed model performs best in Accuracy, Precision, and F1-score when the learning rate is set to 0.01. Besides, the value of Recall is relatively stable despite the learning rate varies from 0.001 to 0.1.
\begin{table*}[htbp]
  \centering
  \caption{The performance of GLAD with different learning rate}
    \begin{tabular}{c|r|r|r|r}
    \toprule
     \diagbox{Metrics}{ Learning Rate}   & 0.001 & 0.005 & 0.01  & 0.1 \\
    \midrule
    Accuracy & 0.872 $\pm$ 0.0162 & 0.8727 $\pm$ 0.0162 & \textbf{0.874} $\pm$ 0.0161 & 0.8727 $\pm$ 0.0162 \\
    \midrule
    Precision & 0.9232 $\pm$ 0.0129 &0.9258 $\pm$ 0.0127& \textbf{0.9279} $\pm$ 0.0125& 0.9248 $\pm$ 0.0128 \\
    \midrule
    Recall & \textbf{0.8115} $\pm$ 0.0190 & 0.8105 $\pm$ 0.0190 & 0.811 $\pm$ 0.0190 & \textbf{0.8115} $\pm$ 0.0190\\
    \midrule
    F1-score    & 0.8638 $\pm$ 0.0166&0.8643 $\pm$ 0.0166 & \textbf{0.8655} $\pm$ 0.0166 & 0.8644 $\pm$ 0.0166\\
    \bottomrule
    \end{tabular}%
  \label{tab:learning rate}%
\end{table*}%

\begin{table}[htbp]
  \centering
  \caption{Number of hidden units in each layer for autoencoder}
    \begin{tabular}{m{1.5cm}<{\centering} m{1.5cm}<{\centering} m{3.5cm} }
    \toprule
    Group&No. layers&No. units\\
    \midrule
     1    &1         &8                       \\
     2    &2         &8,6                        \\
     3    &3         &8,6,4 \\
     4    &4         &8,6,4,2\\
     5    &2         &8,10   \\
     6    &3         &8,10,20 \\
     7    &4         &8,10,20,30  \\
     8    &5         &8,10,20,30,40 \\
    \bottomrule
    \end{tabular}%
  \label{Groups}%
\end{table}%

% Table generated by Excel2LaTeX from sheet 'Sheet6'
\begin{table*}[htbp]
  \centering
  \caption{The performance of GLAD with different hidden layers and hidden units in edge feature learning}
    \begin{tabular}{m{2.5cm}<{\centering} |m{1.5cm}<{\centering}| m{1.5cm}<{\centering} |m{1.5cm}<{\centering} |m{1.5cm}<{\centering} |m{1.5cm}<{\centering} |m{1.5cm}<{\centering} |m{1.5cm}<{\centering} |m{1.5cm}<{\centering}}
    \toprule
   \diagbox{Metrics}{ Group}   & Group 1 & Group 2& Group 3 & Group 4&Group 5&Group 6&Group 7&Group 8\\
   \toprule
    Accuracy & 0.7212 $\pm$ 0.021 & 0.896 $\pm$ 0.0147 & 0.8717 $\pm$ 0.0162 & 0.8777 $\pm$ 0.0156 & \textbf{0.8992} $\pm$ 0.0144 & 0.8945 $\pm$ 0.0146 & 0.8702 $\pm$0.0164 & 0.844 $\pm$ 0.0183 \\
    \midrule
    Precision & 0.6759 $\pm$ 0.0222 & 0.961 $\pm$ 0.0093 & 0.9236 $\pm$ 0.0129& 0.9104 $\pm$ 0.0136 &\textbf{ 0.9645} $\pm$ 0.0089& 0.9433 $\pm$ 0.0110 & 0.9308 +- 0.0124 & 0.9221$\pm$ 0.0135 \\
    \midrule
    Recall & \textbf{0.85} $\pm$ 0.0169 & 0.8255 $\pm$ 0.0183 & 0.8105 $\pm$ 0.0190 & 0.838 $\pm$ 0.0176 & 0.829 $\pm$ 0.0181& 0.8395 $\pm$ 0.0175 & 0.8 +- 0.0196 & 0.7515 $\pm$ 0.02187 \\
    \midrule
    F1-score    & 0.753 $\pm$ 0.0205 & 0.8881 $\pm$ 0.0152 & 0.8634 $\pm$ 0.0167 & 0.8727 $\pm$ 0.0159 & \textbf{0.8916} $\pm$ 0.0149& 0.8884 $\pm$ 0.0150 & 0.8604 $\pm$ 0.0169 & 0.8281 $\pm$ 0.0190 \\
    \bottomrule
    \end{tabular}%
  \label{tab:layers}%
\end{table*}%

\textbf{Number of Hidden Layers.} To investigate the impact of different sizes of neural networks on the performance of GLAD, we set 8 different experimental groups by adjusting the numbers of hidden layers and hidden units in the process of edge feature learning. The detailed settings of the 8 groups can be seen in TABLE~\ref{Groups}. According to TABLE~\ref{tab:layers}, we can conclude that GLAD performs better when the number of hidden layers is set to 2 (Group 5). Furthermore, increasing the number of hidden layers does not always improve the metrics.

\section{Conclusion}
\label{sec:Conclusion and Future Work}
In this work, we aim to solve the problem of anomaly detection in academic citation networks. For this purpose, we develop a novel deep graph learning framework, namely GLAD, to model and identify anomalous citations. We propose to incorporate the node feature learning process and edge feature learning process into the framework. Thus GLAD can make full use of information from node attributes and network structures. Particularly, we propose an effective algorithm to capture semantic information of citation contexts. We conduct comprehensive experiments to validate the performance of GLAD. Experimental results demonstrate the effectiveness of the proposed anomalous citation detection framework as well as the proposed citation purpose classification algorithm.

However, we only consider node attributes and connection information (adjacency matrix) in the node feature learning process. More network structural information such as preferential attachment and Admaic-Adar index could be incorporated. Besides, the construction of anomalous datasets and the labeling of citation purpose are time-consuming. It would be an interesting topic to generate such ground-truth data for further research.

\section*{acknowledgments}
This work is partially supported by National Natural Science Foundation of China under Grant No. 61872054. The authors are grateful to postgraduate students at Dalian University of Technology who have helped with experimental dataset preparation.

\bibliographystyle{IEEEtran}
\bibliography{IEEEabrv,bare_jrnl_ref}

%\newpage

\begin{IEEEbiography}[{\includegraphics[width=1in,height=1.25in,clip,keepaspectratio]{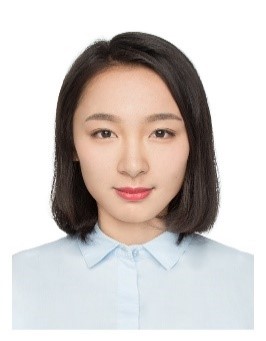}}]{Jiaying Liu}
received the BSc and Ph.D. degrees in software engineering from Dalian University of Technology, China. She is currently an Assistant Professor in School of Economics and Management, Dalian University of Technology, China. Her research interests include data science, big scholarly data, and social network analysis.
\end{IEEEbiography}

\begin{IEEEbiography}[{\includegraphics[width=1in,height=1.25in,clip,keepaspectratio]{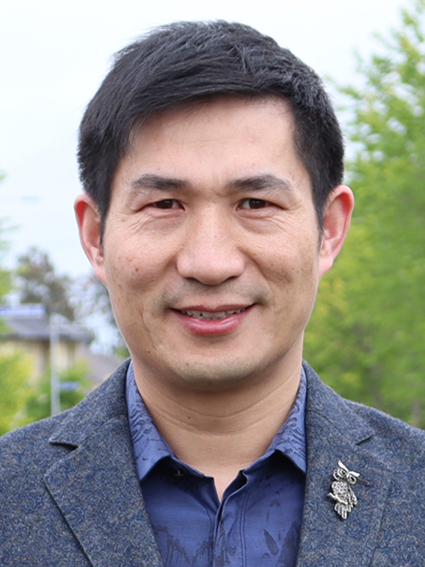}}]{Feng Xia}
(M'07-SM'12) received the BSc and PhD degrees from Zhejiang University, Hangzhou, China. He was Full Professor and Associate Dean (Research) in School of Software, Dalian University of Technology, China. He is Associate Professor and former Discipline Leader (IT) in School of Engineering, IT and Physical Sciences, Federation University Australia. Dr. Xia has published 2 books and over 300 scientific papers in international journals and conferences. His research interests include data science, artificial intelligence, graph learning, and systems engineering. He is a Senior Member of IEEE and ACM.
\end{IEEEbiography}

\begin{IEEEbiography}[{\includegraphics[width=1in,height=1.25in,clip,keepaspectratio]{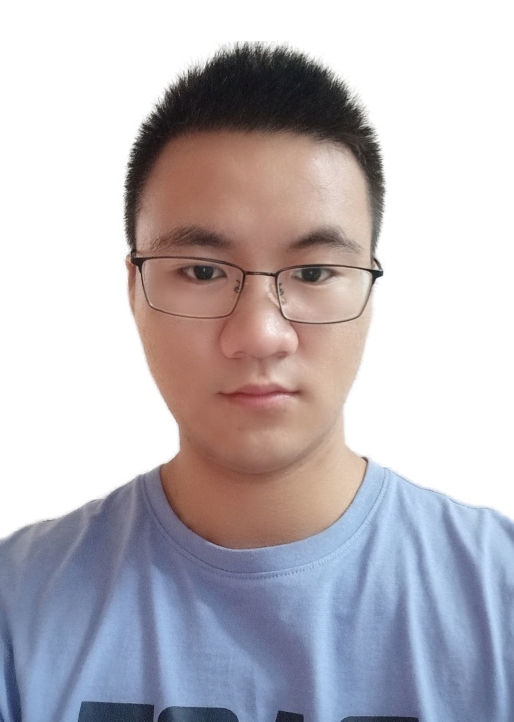}}]{Xu Feng}
received the BSc degree in software engineering from Heilongjiang University, China, in 2019. He is currently pursuing the Master degree in School of Software, Dalian University of Technology, China. His research interests include big scholarly data and data science.
\end{IEEEbiography}

\begin{IEEEbiography}[{\includegraphics[width=1in,height=1.25in,clip,keepaspectratio]{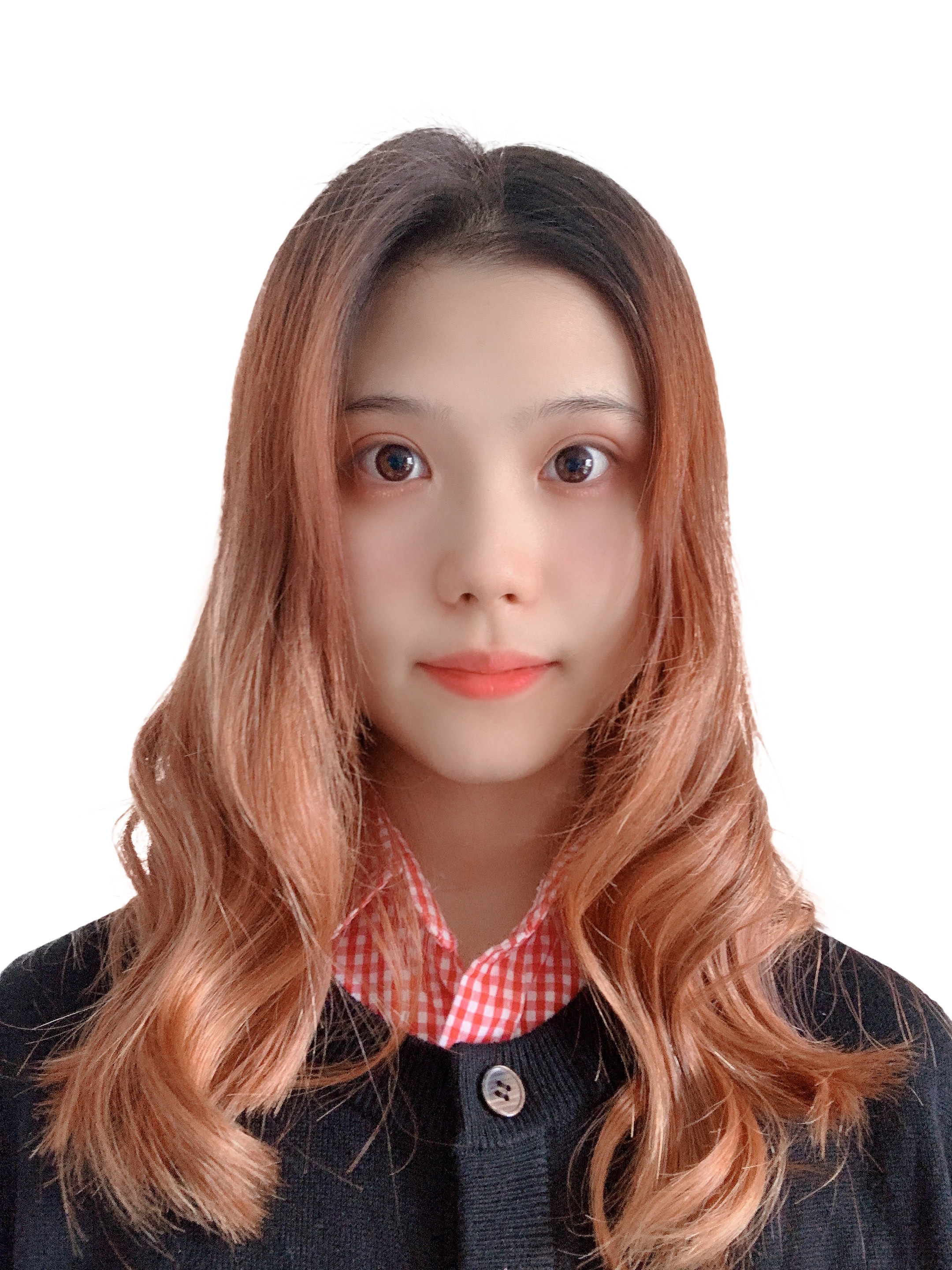}}]{Jing Ren}
received the Bachelor degree from Huaqiao University, China, in 2018, and the Master degree from Dalian University of Technology, China, in 2020. She is currently pursuing the Ph.D. degree in School of Engineering, IT and Physical Sciences, Federation University Australia. Her research interests include data science, graph learning, anomaly detection, and social computing.
\end{IEEEbiography}

\begin{IEEEbiography}[{\includegraphics[width=1in,height=1.25in,clip,keepaspectratio]{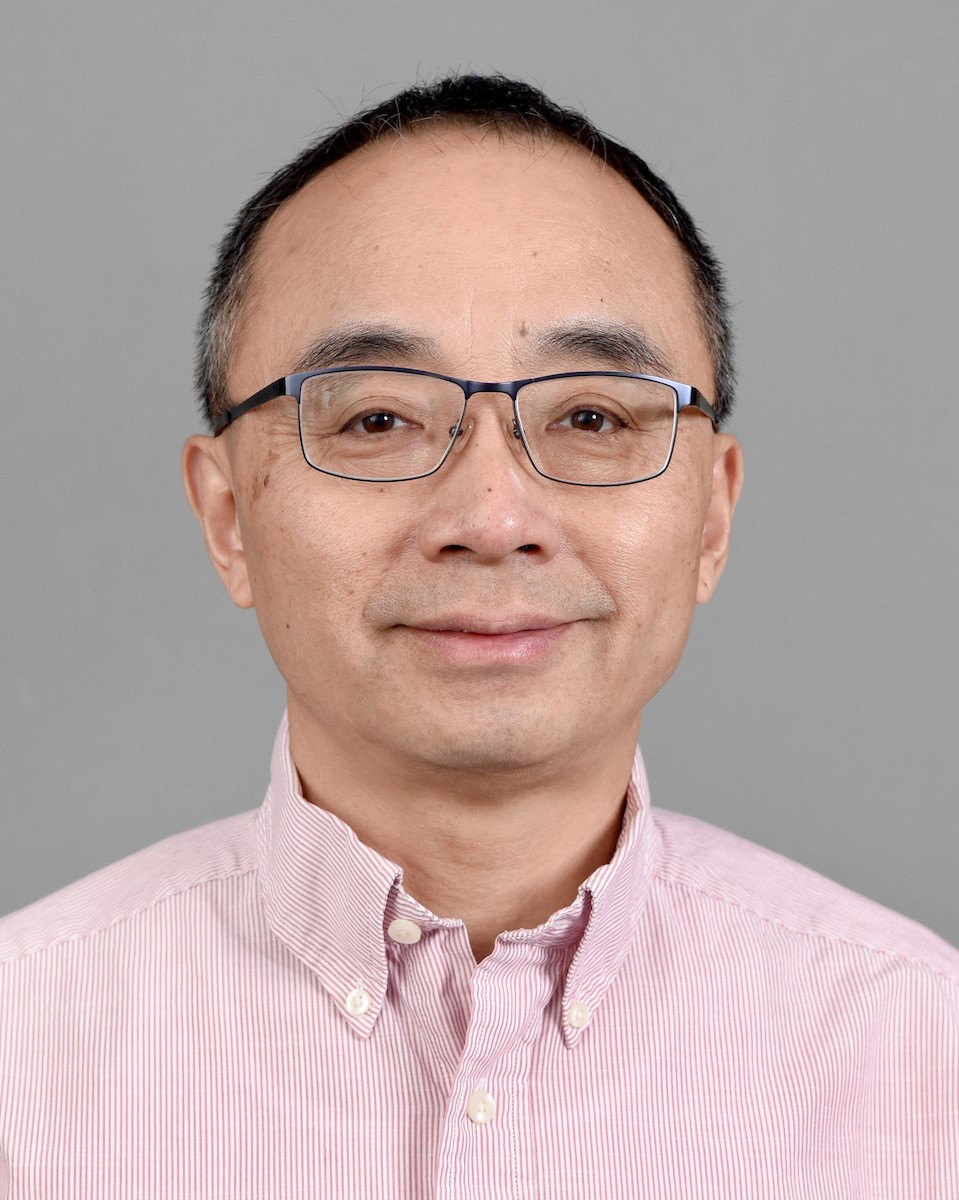}}]{Huan Liu}
(F'12) received the B.Eng. degree in computer science and electrical engineering from Shanghai Jiaotong University and the Ph.D. degree in computer science from the University of Southern California. He is currently a Professor of computer science and engineering at Arizona State University. He was recognized for excellence in teaching and research in computer science and engineering at Arizona State University. His research interests include data mining, machine learning, social computing, and artificial intelligence, investigating problems that arise in many real-world applications with high-dimensional data of disparate forms, such as social media, group interaction and modeling, data preprocessing (feature selection), and text\/web mining. His well-cited publications include books, book chapters, and encyclopedia entries and conference, and journal papers. He serves on journal editorial boards and numerous conference program committees, and is a Founding Organizer of the International Conference Series on Social Computing, Behavioral Cultural Modeling, and Prediction. He is a Fellow of IEEE, ACM, AAAI, and AAAS.
\end{IEEEbiography}

%\vfill

\end{document}